\documentclass[acmsmall]{acmart}
\usepackage{multirow}
\usepackage{graphicx}
\usepackage{subfig}
\AtBeginDocument{%
  }

\setcopyright{acmlicensed}
\copyrightyear{2018}
\acmYear{2018}
\acmDOI{XXXXXXX.XXXXXXX}

\acmJournal{JACM}
\acmVolume{37}
\acmNumber{4}
\acmArticle{111}
\acmMonth{8}

\usepackage{xcolor}

\newcommand{\yankun}[1]{\textcolor{black}{#1}}
\newcommand{\yk}[1]{\textcolor{black}{#1}}
\newcommand{\steven}[1]{\textcolor{black}{#1}}
\newcommand{\stevenNew}[1]{\textcolor{black}{#1}}

\newcommand{\new}[1]{\textcolor{black}{#1}}
\newcommand{\revision}[1]{\textcolor{black}{#1}}

\begin{document}

\title[Modeling Multi-modal Cross-interaction for ML-FSIC Based on Local Feature Selection]{Modeling Multi-modal Cross-interaction for Multi-label Few-shot Image Classification Based on Local Feature Selection}

\author{Kun Yan}
\affiliation{
  \institution{School of Computer Science, Peking University}
  \city{Beijing}
  \country{China}}
\email{kyan2018@pku.edu.cn}

\author{Zied Bouraoui}
\affiliation{
  \institution{CRIL - University of Artois  \& CNRS}
  \country{France}
}
\email{zied.bouraoui@cril.fr}

\author{Fangyun Wei}
\affiliation{
  \institution{School of Computer Science, University of Sydney}
  \country{Australia}
}
\email{fwei8714@uni.sydney.edu.au}

\author{Chang Xu}
\authornote{Corresponding authors.}
\affiliation{
  \institution{School of Computer Science, University of Sydney}
  \country{Australia}
}
\email{c.xu@sydney.edu.au}

\author{Ping Wang}
\authornote{Also with the School of Software and Microelectronics, Peking University and the Key Laboratory of High Confidence Software Technologies (PKU), Ministry of Education.}
\authornotemark[1]
\affiliation{
 \institution{National Engineering Research Center for Software Engineering, Peking University}
 \city{Beijing}
 \country{China}}
\email{pwang@pku.edu.cn}

\author{Shoaib Jameel}
\affiliation{
  \institution{Electronics and Computer Science, University of Southampton}
  \country{UK}}
\email{M.S.Jameel@southampton.ac.uk}

\author{Steven Schockaert}
\affiliation{
  \institution{School of Computer Science and Informatics, Cardiff University}
  \country{UK}}
\email{schockaerts1@cardiff.ac.uk}

\renewcommand{\shortauthors}{Kun et al.}

\begin{abstract}
  The aim of multi-label few-shot image classification (ML-FSIC) is to assign semantic labels to images, in settings where only a small number of training examples are available for each label. A key feature of the multi-label setting is that \revision{an image} often \revision{has} several labels, which typically refer to objects appearing in different regions of the image. When estimating label prototypes, in a metric-based setting, it is thus important to determine which regions are relevant for which labels, but the limited amount of training data and the noisy nature of local features make this highly challenging. As a solution, we propose a strategy in which label prototypes are gradually refined. First, we initialize the prototypes using word embeddings, which allows us to leverage prior knowledge about the meaning of the labels. Second, taking advantage of these initial prototypes, we then use a Loss Change Measurement~(LCM) strategy to select the local features from the training images (i.e.\ the support set) that are most likely to be representative of a given label. Third, we construct the final prototype of the label by aggregating these representative local features using a multi-modal cross-interaction mechanism, which again relies on the initial word embedding-based prototypes. Experiments on COCO, PASCAL VOC, NUS-WIDE, and iMaterialist show that our model substantially improves the current state-of-the-art. 
\end{abstract}




\maketitle

\section{Introduction}
Multi-label image classification (ML-IC) has received considerable attention in recent years \cite{cnn-rnn,chen2019multi,recurrently-dis,orderless}. This task aims to assign descriptive labels to images, where images can have multiple labels. Each label typically describes an object that is shown in the image. As a baseline strategy, we can simply train a binary classifier for each label, to predict the presence of a given type of object. To improve on this strategy, various approaches have been proposed to exploit label dependencies, e.g.\ taking advantage of the fact that the presence of one label makes the presence of another label more (or less) likely \cite{cnn-rnn,orderless,chen2018order,chen2019multi}. In this paper, we are specifically interested in the few-shot setting, i.e.\ the case where only a limited number of training images are available for the labels (e.g., 1 or 5 images per label). Clearly, in this setting, we cannot rely on standard strategies for training binary image classifiers, nor on label co-occurrence statistics.

\begin{figure}[t]
\centering
\includegraphics[width=0.99\columnwidth]{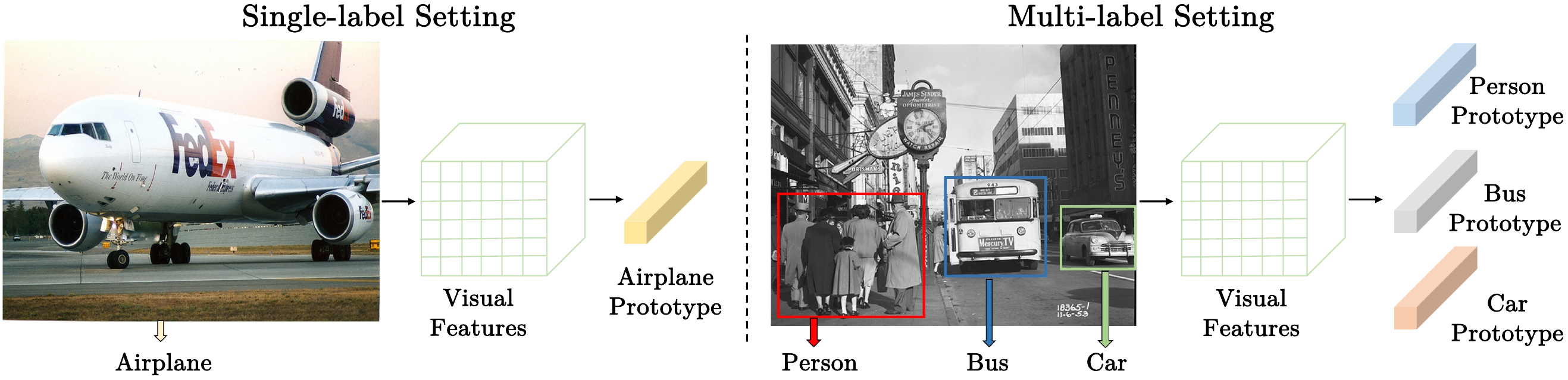} 
\caption{
\steven{In the 1-shot single-label setting, a given training image can be interpreted as a prototype for the considered label (left). In the multi-label setting, labels are related to different regions of the image, and these regions need to be identified before meaningful prototypes can be obtained (right).}
}
\label{fig_sk}
\end{figure}

\yankun{The challenge of few-shot image classification (FSIC) has garnered significant attention, particularly in the single-label (SL) setting~\cite{MAML,Reptile,protonet,match-net,local-mine,dai2023cross,zhang2024fscil}.}
\yankun{Metric-based approaches hold a prominent position in SL-FSIC due to their \steven{strong} performance and \steven{fast} adaptability to new categories. However, it is worth noting that such methods cannot be directly extended to the multi-label setting.}
\yankun{Metric-based approaches usually involve learning} a prototype for each image category and \yankun{subsequently assigning} images to the category whose prototype is closest. 
These prototypes are typically \yankun{derived} by averaging a representation of the training images. In the seminal ProtoNet model \cite{protonet}, for instance, prototypes are simply defined as the average of the global feature maps of the available training examples. This strategy crucially relies on the assumption that most of the image is somehow relevant to its category. In the multi-label setting, such an assumption is highly questionable, given that different labels tend to refer to different parts of the image, as illustrated in Fig.~\ref{fig_sk}. \yankun{For instance, suppose the figure on the right is the only training example with the label \emph{bus}. If we simply use the global feature map of this image as the corresponding prototype, the model is likely to mistakenly assume that other images of cityscapes should also be assigned this label. To obtain a high-quality prototype, it is thus imperative that we can estimate which regions of the image correspond to a given label.}

\yankun{In this paper, we propose a novel} metric-based model for multi-label few-shot image classification (ML-FSIC). Given the aforementioned concerns, we need a strategy that is based on local image features, allowing us to focus on those parts of the training images that are most likely to be relevant. However, as we may only have a single training example for some labels, we cannot implement such a strategy without some kind of prior knowledge about the meaning of the labels. \yankun{Therefore, we} rely on word embeddings, such as those from the GloVe model \cite{glove}, to provide us with this prior knowledge. Some previous works for the single-label setting have already relied on word embeddings for inferring prototypes directly~\cite{am3, aligning}, but as the resulting prototypes are inevitably noisy, they are normally used in combination with prototypes that are derived from visual features. Unfortunately, such visual prototypes are hard to obtain in the ML-FSIC setting. 

For this reason, \yankun{we adopt a different approach to efficiently leverage word embeddings for ML-FSIC. We first construct initial prototypes of the labels based on their word embeddings. Specifically, we introduce a cross-modality loss to learn a joint embedding space, in which the representations of words are aligned with the corresponding visual features.} 
The resulting word representations are too noisy to be used directly.
\yankun{To address this, we propose a multi-modal cross-interaction strategy guided by these noisy prototypes. Concretely, we develop multiple ways of interaction between word embeddings and visual features, including cross-modal attention and cross-modal dynamic convolution, to refine the generation of the final prototype.}
As an important advantage, this strategy can be applied to previously unseen labels without the need for any fine-tuning of the model.
\yankun{We will refer to the model described above as our Base model.}

\yankun{As a further improvement of this base model,}
\yankun{we propose a strategy for identifying the most representative local features prior to prototype generation. To achieve this, we propose a loss change measurement module (LCM), drawing inspiration from \cite{you2019gate}, to improve the selection of local features by assessing which features exert the greatest influence on the loss function.
By filtering noisy local features based on this module, we end up with higher-quality prototypes, but the resulting model has the drawback that it needs to be fine-tuned before it can be applied to unseen labels.}
\yankun{The above-described model will be referred to as the LCM model.}

\yankun{Besides,} we propose a number of changes to the evaluation methodology for ML-FSIC systems. The most important change is concerned with how support sets are sampled, as part of an episode-based strategy. 
The standard $N$-way $K$-shot framework for evaluating FSIC systems is based on the idea that exactly $K$ training images are available for each category of interest. 
While earlier work in ML-FSIC has aimed to mimic this $N$-way $K$-shot framework as closely as possible, we found this to have significant drawbacks when images can have multiple labels. We also propose some changes related to how the query set is sampled and the choice of evaluation metrics. Finally, we propose \yankun{three} new ML-FSIC datasets based on PASCAL VOC~\cite{pascal}, \yankun{NUS-WIDE~\cite{chua2009nus} and iMaterialist~\cite{guo2019imaterialist}}, \steven{by adapting these ML-IC datasets to} the few-shot setting.

\revision{The contributions of this work are summarized as follows:
\begin{itemize}
    \item A novel multi-modal cross-interaction mechanism for prototype generation in ML-FSIC is proposed, featuring channel-wise cross-attention guided by word embeddings and word embedding-based dynamic convolution.
    \item A loss-change measurement module is introduced to identify representative local features, which helps in mitigating the interference of background features.
    \item New ML-FISC benchmarks from NUS-WIDE and iMaterialist are introduced, and extensive experiments on a total of four datasets demonstrate the effectiveness and superiority of our approach over current state-of-the-art methods. 
\end{itemize}}
This paper \revision{is} an extended version of our previous work \cite{yan2021inferring}, with \revision{several key updates.} 
First, we \revision{enhance} the overall approach by introducing a novel multi-modal cross-interaction mechanism, \revision{accompanied by updated experimental results.} 
Second, \revision{we include results for the 5-shot setting and provide a more in-depth analysis.}
Third, we have introduced the Loss Change Measurement (LCM) module, which consistently improves the results.
Our implementation and associated data are available at \url{https://github.com/yk-pku/MMCI-ML-FSIC}.

\section{Related Work}
In this section, we review the related work on \emph{multi-label image classification} (ML-IC), \emph{few-shot image classification} (FSIC), and the combined area of \emph{multi-label few-shot image classification} (ML-FSIC). \steven{We also briefly discuss the impact of Vision-Language models on this field.}

\subsection{Multi-Label Image Classification}
Early solutions for ML-IC simply learned a binary classifier for each label \cite{tsoumakas2007multi}. More recently, various methods have been proposed to improve this basic strategy by exploiting label dependencies in some way. For instance, the CNN-RNN architecture \cite{cnn-rnn} learns a joint embedding space for representing both images and labels, which is used to predict image-label relevance. 
 To avoid the need for a predefined label order, as in RNN-based architectures, \cite{orderless} proposes Minimal Loss Alignment~(MLA) and Predicted Label Alignment~(PLA) to dynamically align the ground truth labels with the predicted label sequence. Some studies~\cite{wang2020multi,chen2019multi,you2020cross,zhou2023double,yuan2023graph} also exploit Graph Convolutional Networks~(GCN)~\cite{gcn} to model label dependencies more explicitly. 
Besides, attention mechanisms play an important role in ML-IC, \steven{as they can} associate labels with specific image regions~\cite{recurrently-dis,spatial-reg,zhu2021residual,zhou2023feature,zhou2023attention}. 
\new{AAMN~\cite{zhou2023attention} proposes an attention-augmented memory network that mines contextual information from various categories within the dataset to enhance input feature representation.} 
\revision{Recently, MMDSR~\cite{kuang2024multi} introduced an approach for dynamically constructing a semantic relationship graph using multi-scale feature reconstruction coupled with channel dual-branch cross-attention.}
 However, the above methods require a large amount of training data and can thus not be directly applied in the few-shot setting.

\subsection{Few-Shot Image Classification}
Different strategies for single-label few-shot image classification have already been proposed, with  metric-based~\cite{relationnet,edge_labeling,few-shot-gnn,feat,cheng2023disentangled} and meta-learning based~\cite{optimization-as-model,MAML,Meta-SGD,meta-latent-embedding} methods being the most prominent. 
Methods of the latter kind use a meta-learner to learn \new{aspects like model weight initialization~\cite{MAML} and update step size~\cite{Meta-SGD}, allowing the adaption of} model parameters to new categories in the few-shot regime. 
It allows adapting architectures to novel tasks with just a few steps of a gradient-based task optimizer. 
However, our method is more closely related to metric-based methods, which aim to learn a generalizable visual embedding space in which different image categories are spatially separated. For instance, ProtoNet~\cite{protonet} generates a visual prototype for each class by simply averaging the embeddings of the support images. The category of a query image is then determined by its Euclidean distance to these prototypes. Instead of using Euclidean distance, the Relation Network~\cite{relationnet} learns to model the distance between query and support images. Other notable models include FEAT~\cite{feat}, which uses a transformer to contextualize the image features relative to the support set, and PSST~\cite{chen2021pareto}, which introduced a self-supervised learning strategy. 

While most metric-based models rely on global features, methods exploiting local features have also been proposed \cite{revisit,zhang2022deepemd}, but these methods are designed for single-label classification. For instance, \cite{revisit} calculates the similarity between all local features of the query image and all local features of the support images. As such, there is no attempt to focus on particular regions of the support images. 
The use of word vectors has also been considered for estimating visual prototypes \cite{am3,aligning,traml,multi-facet}. However, due to the inevitably noisy nature of the predicted prototypes, such methods are best used in combination with prototypes obtained from visual features. 

\subsection{Multi-Label Few-Shot Image Classification}
The ML-FSIC problem has only received limited attention. 
LaSO~\cite{laso} was the first model designed to address this problem. It relies on a data augmentation strategy that generates synthesized feature vectors via label-set operations. 
KGGR~\cite{kggr} uses a GCN to take label dependencies into account, where labels are modelled as nodes, and two nodes are connected if the corresponding labels tend to co-occur. The strength of these label dependencies is normally estimated from co-occurrence statistics, but for labels with limited training data, dependency strength is instead estimated based on GloVe word vectors \cite{glove}. 
\yankun{In \cite{simon2022meta}, several FSIC methods are extended to the multi-label setting, and a neural module is introduced to estimate the label count of a given sample by exploiting relational inference.}
\yankun{BCR~\cite{an2024leveraging} is designed to uncover underlying correlations between instances and labels, incorporating varying levels of importance information. This approach analyzes correlations from both instance-to-label and label-to-instance perspectives.}
In \cite{li2021compositional}, an ML-FSIC method is proposed which learns compositional embeddings based on weak supervision. However, this method is not directly comparable with our method due to its use of weak supervision. 

\yankun{Compared to existing methods such as LaSO~\cite{laso}, KGGR~\cite{kggr}, MetaRE~\cite{simon2022meta}, and BCR~\cite{an2024leveraging}, our work introduces a novel multi-modal cross-interaction mechanism tailored for prototype refinement. This mechanism is based on multiple modes of interaction between representative local features and word embeddings. Our study is the first to comprehensively explore the use of various word embeddings, including GloVe~\cite{glove}, BERT~\cite{DBLP:conf/naacl/DevlinCLT19}, and CLIP~\cite{clip}, specifically for multi-label few-shot image classification.}
\revision{\stevenNew{Several other} studies address multi-label image classification with limited labeled data in different ways, such as leveraging unlabeled data to boost model performance~\cite{li2021semi}, using visual-language models like CLIP to classify images without annotations~\cite{abdelfattah2023cdul}, and training classification models in the presence of ambiguous data~\cite{zhao2021few,zhao2023few}. Nevertheless, our focus is on multi-label few-shot image classification, where only a small number of precisely labeled images (e.g., 1 or 5) are available for each label.}

\subsection{\new{Vision-Language Model}}
\new{Vision-Language (VL) models \steven{use self-supervision strategies to} achieve a strong alignment of textual and visual features, demonstrating an impressive ability to learn generic visual representations~\cite{clip,li2022blip,alayrac2022flamingo,jia2021scaling,yang2023alip,geng2023hiclip,wei2024efficient}. 
\steven{For instance}, CLIP~\cite{clip} is trained with 400 million (image, text) pairs and shows strong zero-shot and transfer capabilities on over 30 visual tasks. With the rise of such powerful vision-language models, various studies have proposed \steven{approaches} to efficiently adapt these models to downstream tasks, such as prompt learning~\cite{zhou2022learning,zhou2022conditional}, and visual adaptation~\cite{gao2024clip,silva2023closer}. 
Although these methods demonstrate remarkable generalization to various zero-shot or few-shot image \steven{classification} tasks, \steven{due to the way they were trained}, they primarily focus on classifying each image into a single label and thus are not well-suited for multi-label settings~\cite{sun2022dualcoop}. 
Recently, TaI-DPT~\cite{guo2023texts} was proposed to extend CLIP \steven{to} multi-label image classification by introducing text-as-image prompting. Although it achieves a 59.2 mAP score on the COCO dataset without \steven{requiring any task-specific} training images, it requires collecting image captions and localized narratives from datasets for training. In contrast, with a single training image per category, our LCM model can achieve a 60.3 mAP score with an extremely shallow backbone Conv-4-64~\cite{simon2022meta,an2024leveraging} and off-the-shelf standard word embedding GloVe~\cite{glove}. 
In the area of multi-label few-shot image classification, our advantage over VL models comes from the ability to achieve promising performance with a \steven{small and thus highly efficient backbone} network. Additionally, our approach is compatible with the backbone of VL models, \steven{offering the potential} for even better performance.
}

\section{Problem Setting}\label{secSetting}
We consider the following multi-label few-shot image classification~(ML-FSIC) setting: we are given a set of base labels $\mathcal{C}_{\textit{base}}$ and a set of novel labels $\mathcal{C}_{\textit{novel}}$, where $\mathcal{C}_{\textit{base}}\cap\mathcal{C}_{\textit{novel}} = \emptyset$. We also have two sets of labelled images: $\mathcal{E}_{\textit{base}}$, containing images with labels from $\mathcal{C}_{\textit{base}}$, and $\mathcal{E}_{\textit{novel}}$, containing images with labels from $\mathcal{C}_{\textit{novel}}$, where $\mathcal{E}_{\textit{base}}\cap\mathcal{E}_{\textit{novel}} = \emptyset$. The images from $\mathcal{E}_{\textit{base}}$ are used for training the model, while those in $\mathcal{E}_{\textit{novel}}$ are used for testing. The goal of ML-FSIC is to obtain a model that performs well for the labels in $\mathcal{C}_{\textit{novel}}$, when given only a few examples of images that have these labels. 
Models are trained and evaluated using so-called episodes. Each training episode involves a support set and a query set. The support set corresponds to the examples that are available for learning to predict the labels in $\mathcal{C}_{\textit{base}}$, while the query set is used to assess how well the system has accomplished this goal.

To construct the support set of a given training episode, for every label in $\mathcal{C}_{\textit{base}}$, we sample $K$ images from $\mathcal{E}_{\textit{base}}$ which have that label, with $K \in \{1, 5\}$. These images are sampled without replacement, meaning that the total number of images in the support set is given by $K \times |\mathcal{C}_{\textit{base}}|$. The query set is sampled in the same way, except that we sample $4$ images per label. Testing episodes are constructed similarly, but with labels from $\mathcal{C}_{\textit{novel}}$ and images from $\mathcal{E}_{\textit{novel}}$ instead.

Note that there are some differences between our sampling strategy and the standard methodology that is adopted for (single-label) FSIC. In particular, FSIC models are usually evaluated using episodes that contain a sub-sample of $N$ classes, where the support set contains exactly $K$ examples of each class. In ML-FSIC, it is difficult to guarantee that the support set contains exactly $K$ examples of each label because most images have multiple labels. The sampling strategy from \cite{laso} nonetheless tries to ensure this, but we argue that doing this has two important drawbacks. 
\yankun{First, adhering to the constraint that each label must occur $K$ times often leads to a limited pool of image combinations available for constructing support sets. This limitation persists even when the requirement is somewhat relaxed. This implies that only a limited number of episodes can be sampled, which in turn makes training the model more challenging and results in less stable test outcomes.}
Second, the total number of images in the support set can vary substantially. For example, if one image contains all labels, then we may have a support set that only contains that one image when $K = 1$. 
\yankun{In contrast, our strategy circumvents these drawbacks. We sample $K$ images for each category without repetition, guaranteeing that at least $K$ examples are accessible for each label. Additionally, the support set of each episode is consistently composed of the same number of images, namely $K \times N$. }
The idea of setting $N = |\mathcal{C}_{\textit{base}}|$ during training and $N=|\mathcal{C}_{\textit{novel}}|$ during testing, rather than selecting a sub-sample of $N$ classes, conforms to the strategy that was used by \cite{laso} and \cite{kggr}. 
\yankun{Furthermore, our approach allows for a larger pool of available episodes. In particular, we sample 200 testing episodes, whereas LaSO only samples 10 testing episodes~\cite{laso}. Moreover, our sampling strategy facilitates episode-based training, closely resembling the evaluation process employed for testing models.}

\section{Method}
\begin{figure}[t]
\centering
\includegraphics[width=1.0\columnwidth]{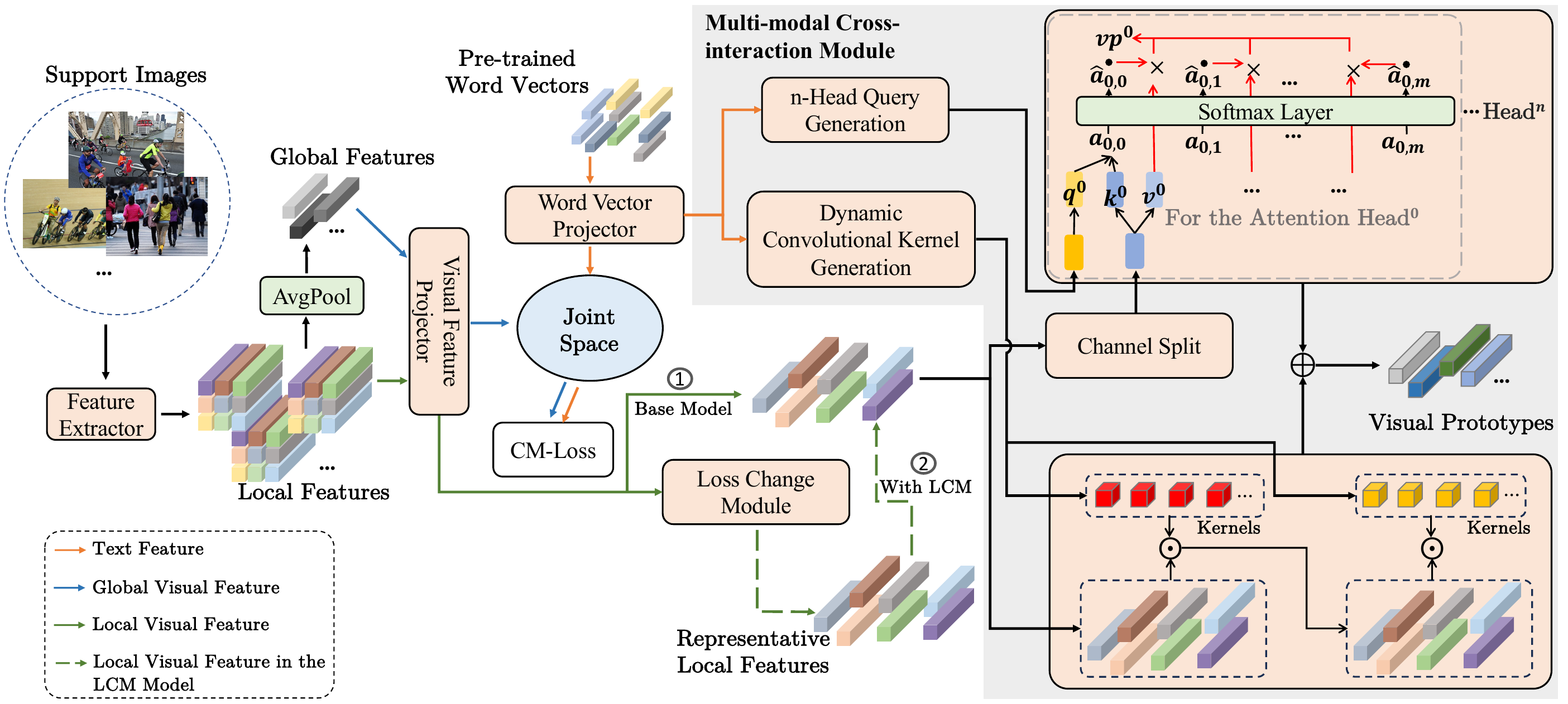} 
\caption{\yankun{Overview of our methodology. }
A joint embedding space is learned in which both labels and images are represented. 
\yankun{A customized multi-modal cross-interaction strategy is proposed to calculate label prototypes using local features from the relevant images in the support set, along with word vectors that provide prior knowledge about the considered label. In our visual representation, the green solid line indicates the local feature flow for the base model, while the green dotted line represents the flow for the LCM model.}}
\label{overview}
\end{figure}

Our base model consists of two main components. The first component \revision{jointly represents label embeddings and visual features in a shared vector space, aiming to predict visual prototypes from the label embeddings.}
Since such prototypes are noisy, \revision{they are not used directly for final label predictions.}
\revision{Instead,} this component is merely used to learn a joint representation of visual features and labels, \revision{as shown in Fig.~\ref{overview}.} 
The second component is aimed at computing the final prototypes by the proposed multi-modal cross-interaction strategy, which mainly consists of a channel-wise cross-attention module and a word embedding-based dynamic convolution module.

In addition to the base model, we also consider a variant, referred to as the LCM model, which involves an additional step between the two components. In particular, in this variant, we introduce Loss Change Measurement (LCM) to make a selection of local features. This selection is also based on the label embeddings from the first module. We then only aggregate the selected local features to construct the final prototypes. The LCM module thus makes a hard selection of the local features, aimed at removing those that are not relevant, before the attention mechanism is used to make a soft selection of the remaining local features.

It is noteworthy that we adopt distinct feature extractors for images and texts for two primary reasons. First, prevalent vision-language models such as CLIP~\cite{clip}, BLIP~\cite{li2022blip} and Flamingo~\cite{alayrac2022flamingo} typically also employ separate extractors for images and texts due to their inherent modality distinctions. Second, \steven{in this way}, our framework can accommodate a variety of image or text extractors, enhancing \steven{its} adaptability.
We now describe these different steps in more detail.

\subsection{Joint Embedding of Visual Features and Labels}\label{jointLearning}
Given an input image $I$, we first use a feature extractor to obtain its local feature map $\mathbf{f}_{\textit{loc}}^{I} \in \mathbb{R}^{n \times h \times w}$, where $n$ is the number of channels, $h$ is the height and $w$ is the width. In this paper, we use a fully convolutional network such as ResNet~\cite{residual} for this purpose. The global visual feature vector $\mathbf{f}_{\textit{glo}}^{I}\in \mathbb{R}^n$ for image $I$ is obtained by averaging the local feature map across the height and width of the image. In particular let us write $f_{i,j,k}$ for the element at position $(i,j,k)$ in the tensor $\mathbf{f}_{\textit{loc}}^{I}$. Then we define $\mathbf{f}_{\textit{glo}}^{I}$ as the vector whose $i\textsuperscript{th}$ component is given by:
\begin{align}\label{eqGlobalImagePlainAverage}
f^*_{i} = \frac{1}{h\cdot w} \sum_{1=j}^h\sum_{1=k}^w f_{i,j,k} .
\end{align}
We use pre-trained word embeddings to represent the labels in $\mathcal{C}_{\textit{base}}\cup \mathcal{C}_{\textit{novel}}$. Let us write $\mathbf{w}_c$ for the word vector representing label $c$, and let $d_w$ be the dimensionality of the word vectors. To represent images and labels in the same vector space, we learn two linear transformations:
\begin{align}
\widehat{\mathbf{f}_{\textit{glo}}^I} &= \mathbf{A}_{\textit{visual}}\, \mathbf{f}_{\textit{glo}}^I  &
\widehat{\mathbf{w}_c} &= \mathbf{A}_{\textit{text}}\, \mathbf{w}_{c} ,
\end{align}
where $\mathbf{A}_{\textit{visual}}\in \mathbb{R}^{d_j \times n}$ is used to project the global feature vector for $I$ onto a space of $d_j$ dimensions. Similarly,  $\mathbf{A}_{\textit{text}}\in \mathbb{R}^{d_j \times d_w}$ is used to project the $d_w$-dimensional embedding of a label $c$ onto the same $d_j$-dimensional space. 
To ensure that the resulting image vectors $\widehat{\mathbf{f}_{\textit{glo}}^I}$ and label representations $\widehat{\mathbf{w}_c}$ are semantically compatible, we propose to use the following loss, which treats the vectors $\widehat{\mathbf{w}_c}$ as prototypes to classify images:
\begin{align} \label{loss_cmw}
	\mathcal{L}_{\textit{cm}} & = - \sum_{I\in \mathcal{S}}\sum_{i=1}^{|\mathcal{C}|} y_{i}^I \cdot \log \sigma(s_{i}^I) + (1 - y_{i}^I) \cdot \log(1 - \sigma(s_{i}^I)) ,
\end{align}
where $\mathcal{S}$ represents the set of images from the support set of the current training episode, $\mathcal{C}=\{c_1,...,c_{|\mathcal{C}|}\}$ is the set of labels, $\sigma(\cdot)$ is the sigmoid function and $y_i^I$ represents the ground truth, i.e.\ $y_i^I=1$ if image $I$ has label $c_i$ and $y_i^I=0$ otherwise. Finally, we have 
\begin{equation}
	s_{i}^I = \lambda\, \cos(\widehat{\mathbf{f}_{\textit{glo}}}^{I}, \widehat{\mathbf{w}_{c_i}}) , 
\label{eq4}
\end{equation}
where $\cos$ denotes the cosine similarity and the scalar $\lambda$ is a hyper-parameter to address the fact that the cosine similarity is bounded between -1 and 1.
Note that we will not be using the vectors $\widehat{\mathbf{w}_c}$ as prototypes for the final classification. We only use this classification loss to align the two modalities (word vectors and visual features). For this reason, we will refer to  Eq.~\eqref{loss_cmw} as the Cross-Modality (CM) loss. 

It should be noted that the proposed alignment process is conceptually straightforward, being based on two standard linear transformations. While more intricate methods may be conceived, we found this approach to achieve satisfactory results. Moreover, linear mappings have proven surprisingly capable for the problem of aligning embedding spaces, often outperforming deep learning based approaches \cite{DBLP:journals/corr/MikolovLS13}.

\subsection{\yankun{Identifying Representative Local Features using Loss Change Measurement}}
\yankun{In our base model, the global image embedding $\mathbf{f}_{\textit{glo}}^{I}$ is obtained by averaging the local features across all regions of the image. However, not all these regions are equally important. Ideally, we only want to include those regions that contain the object being described by a given label, omitting local features that describe the image background. 
In this section, we consider the optional Loss Change Measurement (LCM) module to make a hard selection, aiming to remove those local features that are not relevant. }

\yankun{Let us write $(j,k)$ for the position of some local feature, i.e.\ the local feature corresponding to the region on the $j\textsuperscript{th}$ row and $k\textsuperscript{th}$ column. We now introduce a parameter $\rho_{j,k}\in [0,1]$ to capture the importance of the local feature at position $(j,k)$ of a given image. To learn this parameter, we use the CM loss Eq.~\eqref{loss_cmw}, with two changes. First, the remaining parameters (i.e.\ the matrices $\mathbf{A}_{\textit{visual}}$ and $\mathbf{A}_{\textit{text}}$) are frozen during this step. Second, when computing the global image embedding $\mathbf{f}_{\textit{glo}}^{I}$, we first multiply each local feature by its corresponding importance $\rho_{j,k}$. In particular, the $i\textsuperscript{th}$ component of the vector $\mathbf{f}_{\textit{glo}}^{I}$ is now defined as follows:
\begin{align}\label{eqGlobalWeightedLCM}
f^*_{i} = \frac{1}{h\cdot w} \sum_{1=j}^h\sum_{1=k}^w \rho_{j,k}\,f_{i,j,k} .
\end{align}
Essentially, the importance weight $\rho_{j,k}$ reflects how compatible the local feature at position $(j,k)$ is with the relevant label embeddings (i.e.\ the embeddings of the labels assigned to the image). After each iteration, the importance weights are normalised, to ensure that $\rho_{j,k}\in [0,1]$. In particular, we update each $\rho_{j,k}$ as follows:
\begin{align}
\rho_{j,k} := \frac{\rho - \rho_{\textit{min}}}{\rho_{\textit{max}}-\rho_{\textit{min}}} ,
\end{align}
with
\begin{align}
\rho_{\textit{min}} = \min_{1\leq j\leq h} (\min_{1\leq k\leq w} (\rho_{j,k})) \ , \quad  
\rho_{\textit{max}} = \max_{1\leq j\leq h} (\max_{1\leq k\leq w} (\rho_{j,k}))
\end{align}
If we end up with $\rho_{j,k}=0$, this would mean that the local feature $(j,k)$ is permanently removed. To prevent this from happening, in such a case, we instead set the value of $\rho_{j,k}$ as the lowest non-zero value of the importance weight $\rho_{j',k'}$, across all local features.}

\yankun{Rather than directly using the weights $\rho_{j,k}$ to make the selection of local features, taking inspiration from \cite{you2019gate,local-mine}, we measure the impact on the classification loss when a given feature is (partially) removed. Intuitively, if removing a given local feature has a large impact on the classification loss, it is likely to be representative of the object we are trying to model (i.e.\ the object described by the target label). In contrast, if the impact is minimal, it is more likely that this feature describes the image background or a different object.
The impact of removing the local feature $(j,k)$ on the loss function is computed as follows:
\begin{equation} \label{loss_change}
g(j,k) =  |\mathcal{L}^{(j,k)}_{\Omega}(\rho_{j,k}) - \mathcal{L}^{(j,k)}_{\Omega}(0)|,
\end{equation}
where $\mathcal{L}_{\Omega}$ represents a classification loss with parameters $\Omega$.  We again use the CM loss $\mathcal{L}_{\textit{cm}}$ as the classification loss, where we assume that $\mathbf{f}_{\textit{glo}}^{I}$ is computed based on the importance weights, according to Eq.~\eqref{eqGlobalWeightedLCM}. The argument of $\mathcal{L}^{(j,k)}_{\Omega}$ represents the importance weight of the local feature at position $(j,k)$. Note that $\mathcal{L}^{(j,k)}_{\Omega}(0)$ thus represents the value of the loss function we obtain after setting $\rho_{j,k}$ to 0, i.e.\ if we completely disregard the region of the image at position $(j,k)$. In other words, $g(j,k)$ captures the difference in the loss function between (i) using $\rho_{j,k}$ as the importance of the local feature at position $(j,k)$ and (ii) setting this weight to 0. Intuitively, if $g(j,k)$ is above some threshold (for the value of $\rho_{j,k}$ upon convergence), we keep the local feature at position $(j,k)$; otherwise, we remove it. However, in practice, we do not use $g(j,k)$ directly. In particular, (i) we use an approximation of $g(j,k)$ to make the computation more efficient and (ii) rather than only evaluating $g(j,k)$ after the value of $\rho_{j,k}$ has converged, we use a momentum based strategy to improve stability. We now explain these two steps in more detail.}

\yankun{To simplify the implementation, we take advantage of the Taylor approximation of $\mathcal{L}^{j,k}_{\Omega}(x)$ at $x_{0}$:
\begin{align} 
	\mathcal{L}^{j,k}_{\Omega}(x) = \mathcal{L}^{j,k}_{\Omega}(x_{0}) + \frac{\mathcal{L}^{{j,k}(1)}_{\Omega}(x_{0})}{1!}(x-x_{0}) + ...+ \frac{\mathcal{L}^{{j,k}(n)}_{\Omega}(x_{0})}{n!}(x-x_{0})^{n} + R_{n}(x) ,
\end{align}
where we write $\mathcal{L}^{{j,k}(i)}_{\Omega}$ for the $i\textsuperscript{th}$ derivative of  $\mathcal{L}^{^{j,k}}_{\Omega}$.
We find that $\mathcal{L}^{j,k}_{\Omega}(0)$ can be estimated as $\mathcal{L}^{j,k}_{\Omega}(\rho) - \rho\mathcal{L}^{{j,k}(1)}_{\Omega}(\rho)$ by setting $x=0$ and $x_{0} = \rho$. Therefore, Eq.~\ref{loss_change} can be estimated as: 
\begin{equation}\label{eqLCMapproximation}
	g(j,k) \approx |\rho_{j,k}\mathcal{L}^{j,k(1)}_{\Omega}(\rho_{j,k})| .
\end{equation}
Note that the derivatives $\mathcal{L}^{j,k(1)}_{\Omega}(\rho_{j,k})$ can be computed in parallel for all local features, which makes using this approximation much faster than evaluating Eq.~\eqref{loss_change} for every local feature.}

\yankun{As already mentioned, the importance weights $\rho_{j,k}$ are iteratively learned by optimising the CM loss. These weights are initialized as $\rho_{j,k}=1$. Rather than only using the converged values of $\rho_{j,k}$, we use a momentum-based strategy to improve the stability of the estimation. In particular, drawing on the experience of \cite{tarvainen2017mean}, we compute the following estimate in iteration $i$:
\begin{align}
	   f_{i}(j,k) = \alpha_{i}\,f_{i - 1}(j,k) + (1 - \alpha_{i}) \,g_{i}(j,k) \ , \quad   
      \alpha_{i} = \textit{min}\left(1 - \frac{1}{({i + 1})}, \alpha\right), 
\end{align}
where we set $f_{0}(\rho) = 0$ and $\alpha = 0.95$ following \cite{tarvainen2017mean}. The value of $g_{i}(j,k)$ is computed according to Eq.~\eqref{eqLCMapproximation}, using the importance weights $\rho_{j,k}$ that were obtained in the previous iteration.}

\yankun{After the final iteration $n$, we make the following decision: the local feature at position $(j,k)$ is retained if $\sigma(f_n(j,k)) \geq \theta$, where $\sigma(\cdot)$ denotes the sigmoid function.
Otherwise, this local feature is removed. }

\subsection{\yankun{Multi-modal Cross-interaction for Prototype Construction}}\label{proto_cons}
We now explain how the final label prototypes are constructed. \yankun{The multi-modal cross-interaction module, depicted in Fig.~\ref{overview}, \steven{allows} two distinct \steven{types} of information interaction between visual and text features. The first is to aggregate various visual patterns using word vectors as \emph{query} within a customized cross-attention mechanism. The second is to employ word embeddings to generate convolutional kernels that interact with image features.} 
\subsubsection{\yankun{Channel-\steven{wise} Cross-attention Guided by Word Embedding}}
\yankun{As pointed out in~\cite{chu2016structured}, features learned from different channels correspond to distinct visual patterns. Utilizing the word embedding-based prototypes from Sec.~\ref{jointLearning} as prior knowledge, we extract various visual patterns from local features.}
\yankun{Concretely, consider a label $c$ and suppose this label has been assigned to $m$ images from the support set. \steven{We} can derive a total of $l= h \cdot w \cdot m$ local feature vectors from these $m$ images. Let us write these local feature vectors as $\mathbf{u}_1,...,\mathbf{u}_l$. These feature vectors are first projected into the joint embedding space: $\widehat{\mathbf{u}_i} = \mathbf{A}_{\textit{visual}}\, \mathbf{u}_i$. Subsequently, \steven{the} feature vectors are split evenly along the channel dimension to serve as \emph{value} and \emph{key} inputs for $n_a$ attention heads within the attention mechanism~\cite{attention}. The corresponding label embedding $\widehat{\mathbf{w}_c}$ provides the \emph{query} for each attention head through a fully connected (FC) layer.} Specifically, we have:
\yankun{\begin{align}
\mathbf{q}^j_c = \mathbf{Q}_j \widehat{\mathbf{w}_c} & \ , \quad 
\mathbf{k}^j_i = \mathbf{v}^j_i = \widehat{\mathbf{u}_i}[(j-1) \cdot d_a:j \cdot d_a] , \\
\mathbf{p}_c^j = \sum_{i=1}^{l} \mu_i^j \mathbf{v}^j_i & \ , \quad 
(\mu^j_1,...,\mu^j_l) = \textit{Softmax}\left(\frac{\mathbf{q}^j_c\cdot \mathbf{k}^j_1} {\sqrt{d_a}},...,\frac{\mathbf{q}^j_c\cdot \mathbf{k}^j_l}{\sqrt{d_a}}\right) 
\end{align}
where $d_a = \frac{d_j}{n_a} $ is the dimensionality of the vectors $\mathbf{q}^j_c$, $\mathbf{k}^j_i$, and $\mathbf{v}^j_i$.
The vector $\mathbf{p}_c^j$ represents the contribution of the j\textsuperscript{th} attention head to the prototype of label $c$.}
\yankun{As there are $n_a$ attention heads in total, the prototype of label $c$ in this cross-attention mechanism is given by:}
\begin{align}
\mathbf{p}_{c}^{att} = \textit{MLP}(\mathbf{p}_c^1 \oplus ... \oplus \mathbf{p}_c^{n_a}) ,
\label{eqPrototypeComputation}
\end{align}
where we write $\oplus$ for vector concatenation and \textit{MLP} consists of two linear layers with GeLU activation and dropout.

\subsubsection{\yankun{Word Embedding-based Dynamic Convolution}}
\yankun{Inspired by the use of dynamic convolution in~\cite{sun2021sparse} to enhance object features, we propose generating convolutional kernels from word embeddings to interact with image features. Specifically, we begin by selecting the top $n_d$ local features from the joint embedding space, denoted as $\widehat{\mathbf{u}_1}, \dots, \widehat{\mathbf{u}_{n_d}}$, based on their cosine similarities to $\widehat{\mathbf{w}_c}$. We then utilize linear layers to produce two sets of convolutional kernels from $\widehat{\mathbf{w}_c}$, i.e. $\Theta_{c}^{1} \in \mathbb{R}^{1 \times 1 \times d_{j} \times d_{c}}$ and $\Theta_{c}^{2} \in \mathbb{R}^{1 \times 1 \times d_{c} \times d_{j}}$. Finally, the operation of word embedding-based dynamic convolution is defined as :}
\yankun{\begin{align}
	\widebar{\mathbf{u}}_{i} &= \textit{ReLU}(\textit{Norm}(\Theta_{c}^{1} \widehat{\mathbf{u}_{i}})), \\
	\mathbf{p}_{c}^{dyn} &= \frac{1}{n_d} \sum_{i=1}^{n_d} \textit{ReLU}(\textit{Norm}(\Theta_{c}^{2} \widebar{\mathbf{u}}_{i})),
\end{align}}
\yankun{where $\textit{ReLU}$ and $\textit{Norm}$ denote the ReLU activate function and Layer Normalization~\cite{sun2021sparse}, respectively.}
\yankun{Finally, the prototype of category $c$ is obtained as:}
\yankun{\begin{align}
	\mathbf{p}_{c} = \mathbf{p}_{c}^{att} + \mathbf{p}_{c}^{dyn}.
\end{align}}

\subsubsection{\yankun{Loss Function}}
\yankun{To train our multi-modal cross-interaction module, we use the following loss function, which we refer to as the Query Loss:
\begin{align}
	\mathcal{L}_{\textit{query}} & = - \sum_{I\in\mathcal{Q}}\sum_{i=1}^{|\mathcal{C}|} y_{i}^I \cdot \log \sigma(q_{i}^I) + (1 - y_{i}^I) \cdot \log(1 - \sigma(q_{i}^I)) , 
\end{align}
where $\mathcal{Q}$ represents the set of images from the query set of the current training episode. As before, $\mathcal{C}$ is the set of labels and $y_i^I$ represents the ground truth. The predictions $q_i$ are obtained as follows:
\begin{align}
	q_{i}^I = \lambda\cos(\widehat{\mathbf{f}_{\textit{glo}}}^{I}, \mathbf{p}_{c_{i}}) , 
\label{q_p}
\end{align}
with $\lambda$ the same scalar as in Eq.~\eqref{eq4}. Here we assume that the global feature vector $\widehat{\mathbf{f}_{\textit{glo}}}^{I}$ is computed as a standard average, according to Eq.~\eqref{eqGlobalImagePlainAverage}, as we are not able to compute importance weights for query images. As an alternative to using the global feature vector $\widehat{\mathbf{f}_{\textit{glo}}}^{I}$ we could also compare $\mathbf{p}_{c_{i}}$ to the local image features. However, as we will see in our analysis in Section \ref{secExperiments}, this variant does not lead to clear improvements while having a considerably higher computational cost. Note that there are two key differences between $\mathcal{L}_{\textit{cm}}$ and $\mathcal{L}_{\textit{query}}$: (i) $\mathcal{L}_{\textit{cm}}$ is trained using the support images while $\mathcal{L}_{\textit{query}}$ is trained using the query images; and (ii) prototypes in $\mathcal{L}_{\textit{cm}}$ are estimated from word vectors while prototypes in $\mathcal{L}_{\textit{query}}$ are those obtained by aggregating local visual features.}

\subsection{Model Training and Evaluation}
The model is trained by repeatedly sampling training episodes from $\mathcal{C}_{\textit{base}}$, as explained in Section \ref{secSetting}. Given a training episode with support set $\mathcal{S}$ and query set $\mathcal{Q}$, the model parameters are updated using the following loss:
\begin{align}
    \mathcal{L}_{\textit{all}} = \mathcal{L}_{\textit{cm}} + \gamma \mathcal{L}_{\textit{query}} , 
\end{align}
with $\gamma$ a hyperparameter to control the relative importance of both components. 

After the model has been trained, it can be evaluated on test episodes as follows. 
For the base model, without the LCM step, we first construct the prototypes using the support set, as in Eq.~\eqref{eqPrototypeComputation}. Note that we can do this without fine-tuning any model parameters. For each query image $I$, the probability that it has label $c_i$ is computed as $\sigma(q_i^I)$ with $q_i^I$ as defined in Eq.~\eqref{q_p}. For the variant with LCM, we first need to use the Loss Measurement Module to identify the most representative local features in the images in the support set. The labels are then predicted in the same way as for the base model.

\section{Experiments}\label{secExperiments}
\subsection{Experimental Setup}
\subsubsection{Datasets}
We have conducted experiments on four datasets, starting with COCO~\cite{lin2014microsoft}. This dataset was already used for evaluating LaSO in \cite{laso}, where a split into 64 training labels and 16 test labels was proposed. However, as this split did not include a validation set, we split their 64 training labels into 12 labels for validation (\textit{cow}, \textit{dining table}, \textit{zebra}, \textit{sandwich}, \textit{bear}, \textit{toaster}, \textit{person}, \textit{laptop}, \textit{bed}, \textit{teddy bear}, \textit{baseball bat}, \textit{skis}) and 52 labels for training, while keeping the same 16 labels for testing. We follow \cite{laso} to include images from the COCO 2014 training and validation sets. The images which do not contain any of the test and validation labels are used as the training set. Similarly, the validation set only contains images that do not contain any training or test labels. 
Second, we propose a new ML-FSIC dataset based on PASCAL VOC \cite{pascal}, which has 20 labels. To use as many images as possible, we select the following six labels for the novel set $\mathcal{C}_{\textit{novel}}$: \textit{dog}, \textit{sofa}, \textit{cat}, \textit{potted plant}, \textit{tv monitor}, \textit{sheep}. The following six labels were selected for the validation split: \textit{boat}, \textit{cow}, \textit{train}, \textit{aeroplane}, \textit{bus}, \textit{bird}. The remaining eight labels are used for training. We use the images from the VOC 2007 training, validation, and test splits, as well as the VOC 2012 training and validation splits (noting that the labels of the VOC 2012 test split are not publicly available). 
\yankun{Third, we propose a new ML-FSIC dataset based on NUS-WIDE~\cite{nus-wide}. NUS-WIDE contains 161789 images for training and 107859 images for testing, which were manually annotated with 81 concepts. In our setting, we use both training and testing images. We have selected 17 labels for the novel set (\textit{bear}, \textit{cityscape}, \textit{statue}, \textit{bridge}, \textit{sunset}, \textit{sign}, \textit{coral}, \textit{fish}, \textit{fire}, \textit{cars}, \textit{sand}, \textit{vehicle}, \textit{police},\textit{temple}, \textit{horses}, \textit{reflection}, \textit{street}) and 12 labels for the validation set (\textit{tower}, \textit{tree}, \textit{food}, \textit{sun}, \textit{town}, \textit{beach}, \textit{military}, \textit{birds}, \textit{flowers}, \textit{house}, \textit{lake}, \textit{garden}). The training set consists of the remaining labels. The complete NUS-WIDE dataset is no longer available from the official website\footnote{\url{https://lms.comp.nus.edu.sg/wp-content/uploads/2019/research/nuswide/NUS-WIDE.html}}, as many image download links have expired. Therefore, we have instead collected the images from the NUS-WIDE dataset from a Kaggle website\footnote{\url{https://www.kaggle.com/datasets/xinleili/nuswide?resource=downloadx}}.}
\yankun{Lastly, we introduce a new ML-FSIC dataset based on iMaterialist~\cite{guo2019imaterialist}. Following a similar methodology as MoFSOD~\cite{lee2022rethinking}, we extract 46 classes. Out of these, 16 labels (\textit{sock}, \textit{jumpsuit}, \textit{glove}, \textit{scarf}, \textit{coat}, \textit{belt}, \textit{zipper}, \textit{cardigan}, \textit{jacket}, \textit{hood}, \textit{vest}, \textit{flower}, \textit{tie}, \textit{collar}, \textit{ruffle}, \textit{shirt\_blouse}) are selected for the novel set, while 12 labels (\textit{watch}, \textit{lapel}, \textit{shorts}, \textit{bead}, \textit{glasses}, \textit{hat}, \textit{sweater}, \textit{tights\_stockings}, \textit{buckle}, \textit{skirt}, \textit{applique}, \textit{rivet}) are allocated \steven{to} the validation set. The remaining labels constitute the training set.} 
\yankun{An overview of the labels contained in the four datasets can be found in the Appendix.}

\subsubsection{Methodology}
\label{train+test}
Every model is trained for 200 epochs, with the first 10 epochs used as a warm-up. We used the Adam optimizer with an initial learning rate of 0.001. \yankun{The variant with LCM requires a training step during the test phase, to learn the importance weights of the local features of the support images.  We set the number of epochs for this training step to 20. For baseline methods that require a training or fine-tuning step, we set the number of epochs to 40 (noting that 20 epochs are not sufficient for some of these methods to converge). Note that our base model is used without any training or fine-tuning.}
During the test phase, we sample 200 test episodes. The results are averaged over all test episodes. \yankun{We report results in terms of macro and micro metrics. The macro metric is evaluated by averaging per-class predictions while the micro metric is an overall measure for all images over all labels. Specifically, we use macro/micro average precision (Ma-AP/Mi-AP) and macro/micro F1-measure (Ma-F1/Mi-F1), following earlier work in ML-IC~\cite{spatial-reg}. Ma-AP is also denoted as mAP in many works~\cite{laso,cole2021multi,durand2019learning}.}
We assume that a label is predicted as positive if its estimated probability is greater than 0.5 following the standard setting~\cite{spatial-reg}.

\subsubsection{Implementation Details}\label{exp:imple-details}
We use ResNet-50~\cite{residual}, \yankun{ViT~\cite{vit} and GoogleNet-v3~\cite{szegedy2016rethinking}} as feature extractors. The dimensionality of the joint embedding space was set as $d_j=512$. 
As word embeddings, we mainly consider 300-dimensional GloVe \cite{glove} vectors\footnote{Specifically, we use vectors that were trained from Wikipedia 2014 and Gigaword 5, which we obtained from the GloVe project page, at \url{https://nlp.stanford.edu/projects/glove}.} as this is the most common choice in the literature for few-shot image classification~\cite{traml,am3} and multi-label image classification~\cite{kggr,chen2019multi}. In addition, we also experimented with the standard pre-trained FastText and Word2Vec embeddings from an online repository\footnote{https://developer.syn.co.in/tutorial/bot/oscova/pretrained-vectors.html}. 
\yankun{Beyond standard word embeddings, we also experimented with embeddings that were extracted from BERT \cite{DBLP:conf/naacl/DevlinCLT19} using a number of different strategies. In particular, we use mirrorBERT \cite{mirrorbert} and the biencoder model from \cite{biencoder} (referred to as BERT\textsubscript{biencoder}), both of which are fine-tuned BERT models that can map any word or phrase onto a vector. In addition, we also use mirrorWiC \cite{mirrorwic}, and the strategies from \cite{DBLP:conf/ijcai/LiBCAGS21} (referred to as BERT\textsubscript{avg}) and \cite{concn} (referred to as BERT\textsubscript{concn}), all of which learn embeddings of words in context. To use these strategies to learn the embedding of a given word, we randomly select 500 sentences that mention the word from Wikipedia. Then we obtain the contextualised representations of these word mentions, using the aforementioned models, and finally compute the average of these vectors across the 500 sentences. Finally, we also consider word vectors extracted from CLIP~\cite{clip}, using ``a photo of a [CLASS]'' as the prompt to guide its text encoder.}
Based on the validation split, for both datasets, the number of attention heads was set to $8$, 
the $\theta$ in LCM was set to $0.65$, and $\gamma$ was set to $1$. We present our results for two variants: the base model and the LCM model. 
Note that an important advantage of the base model is its ability to classify novel categories without requiring any parameter adjustments. On the other hand, using the LCM module should lead to better results.

\begin{table*}[t]
\caption{
Overview of the main results on COCO, in terms of micro/macro AP and F1 scores (in $\%$).
The best results are highlighted in bold font. ``-'' indicates that no word embeddings are used.
}
\centering
\footnotesize
\begin{tabular}{l c cccccccc}
\toprule
 & \multirow{2}{*}{Word-Emb.} & \multicolumn{4}{c}{\textbf{1-shot}} & \multicolumn{4}{c}{\textbf{5-shot}}\\
\cmidrule(l){3-6} \cmidrule(l){7-10}
& & Mi-AP & Mi-F1 & Ma-AP & Ma-F1                & Mi-AP & Mi-F1 & Ma-AP & Ma-F1 \\
\midrule
ResNet-50   & - & 11.31 & 12.51 & 19.67 & 13.27    & 19.12 & 16.08 & 24.17 & 17.22 \\ 
ResNet-101  & - & 10.22 & 12.04 & 18.87 & 12.18    & 17.93 & 16.32 & 21.67 & 15.74 \\ 
ViT         & - & 10.07 & 11.02 & 16.94 & 11.13    & 24.73 & 21.10 & 28.78 & 20.93 \\ 
ResNet-50 + ViT    & - & 10.13 & 10.21 & 16.71  & 11.17    & 23.65  & 20.41 & 26.65 & 19.56 \\ 
PLA                & Self-learned & 21.33 & 20.89 & 30.61  & 20.96    & 27.87  & 29.91 & 36.38 & 28.25 \\ 
PLA~(GloVe)        & GloVe   & 23.02 & 22.15 & 30.13  & 20.37    & 27.11  & 28.48 & 35.13 & 27.47   \\ 
LaSO               & -   & 16.83 & 15.09 & 25.41  & 16.11    & 22.24  & 20.14 & 28.41 & 21.59 \\ 
MAML               & - & 25.42 & 23.11 & 35.30  & 23.82    & 28.30  & 30.50 & 37.89 & 29.13 \\
\revision{WGA}   & \revision{GloVe} & \revision{35.30} & \revision{34.21} & \revision{42.84}  & \revision{28.91} 
                 & \revision{38.97}  & \revision{38.10} & \revision{44.91} & \revision{33.12} \\
\midrule
Base model       & GloVe & \yankun{39.68} & \yankun{38.24}  & \yankun{47.02}  & \yankun{32.94}  & \yankun{42.84} & \yankun{42.17} & \yankun{48.85} & \yankun{37.47}  \\ 
LCM model        & GloVe & \yankun{41.02} & \yankun{38.67}  & \yankun{48.92}  & \yankun{33.55}   & \yankun{46.35} & \yankun{44.86} & \yankun{52.79} & \yankun{39.01} \\ 
\midrule
Base model       & mirrorBERT & \yankun{41.79} & \yankun{39.42}  & \yankun{49.49} & \yankun{34.46}  & \yankun{45.15} & \yankun{43.26} & \yankun{51.49} & \yankun{39.54}  \\ 
LCM model        & mirrorBERT & \textbf{\yankun{43.20}} & \textbf{\yankun{40.39}} & \textbf{\yankun{51.55}} & \textbf{\yankun{35.55}}
                       & \textbf{\yankun{47.72}} & \textbf{\yankun{45.81}} & \textbf{\yankun{54.98}} & \textbf{\yankun{42.04}} \\ 
\bottomrule
\end{tabular}
\label{coco}
\end{table*}

\begin{table*}[h]
\caption{
\revision{Overview of the main results on COCO, in terms of micro/macro AP and F1 scores (in $\%$), following the same data split as in FsPML~\cite{zhao2023few}.
The best results are highlighted in bold font. ``-'' indicates that no word embeddings are used.}
}
\centering
\footnotesize
\begin{tabular}{l c cccccccc}
\toprule
 & \multirow{2}{*}{Word-Emb.} & \multicolumn{4}{c}{\textbf{1-shot}} & \multicolumn{4}{c}{\textbf{5-shot}}\\
\cmidrule(l){3-6} \cmidrule(l){7-10}
& & Mi-AP & Mi-F1 & Ma-AP & Ma-F1                & Mi-AP & Mi-F1 & Ma-AP & Ma-F1 \\
\midrule
ResNet-50   & - & 18.43 & 18.38 & 25.03 & 18.92    & 26.53 & 24.62 & 32.91 & 25.33 \\ 
ResNet-101  & - & 15.85 & 17.93 & 24.61 & 18.77    & 23.79 & 23.38 & 28.37 & 22.98 \\ 
ViT         & - & 14.53 & 16.29 & 21.00 & 17.36    & 30.46 & 26.77 & 35.83 & 26.41 \\ 
ResNet-50 + ViT    & - & 14.44 & 15.18 & 22.56  & 18.28    & 31.02  & 26.58 & 34.61 & 26.22 \\ 
PLA                & Self-learned & 29.34 & 29.01 & 37.68  & 28.30    & 35.84  & 37.17 & 43.75 & 36.71 \\ 
PLA~(GloVe)        & GloVe   & 30.44 & 29.16 & 37.81  & 27.56    & 35.06  & 36.61 & 42.14 & 35.02   \\ 
LaSO               & -   & 24.75 & 22.21 & 33.02  & 23.81    & 29.59  & 27.62 & 35.47 & 29.35 \\ 
MAML               & -   & 33.02 & 29.37 & 42.03  & 29.84    & 35.91  & 37.72 & 44.78 & 36.05 \\
WGA                & GloVe   & 39.26 & 37.43 & 47.25  & 33.60    & 42.71  & 43.04 & 49.55 & 39.01 \\
\midrule
Base model       & GloVe & 46.44 & 44.82  & 52.86  & 39.21   & 49.68 & 48.76 & 55.21 & 44.75  \\ 
LCM model        & GloVe & 47.37 & 45.17  & 54.78  & 40.57   & 52.99 & 50.76 & 58.72 & 45.92 \\ 
\midrule
Base model       & mirrorBERT & 48.81 & 45.60  & 55.94 & 40.83  & 51.08 & 49.43 & 57.90 & 45.43  \\ 
LCM model        & mirrorBERT & \textbf{50.06} & \textbf{47.35} & \textbf{56.75} & \textbf{41.02}
                       & \textbf{53.92} & \textbf{51.84} & \textbf{60.19} & \textbf{47.85} \\ 
\bottomrule
\end{tabular}
\label{coco_fspml}
\end{table*}

\begin{table*}[t]
\caption{\revision{Overview of the main results on NUS-WIDE, in terms of micro/macro AP and F1 scores (in $\%$), following the same data split as in FsPML~\cite{zhao2023few}. The best results are highlighted in bold font. ``-'' indicates that no word embeddings are used.}}
\centering
\footnotesize
\begin{tabular}{l c cccccccc}
\toprule
  & \multirow{2}{*}{Word-Emb.} & \multicolumn{4}{c}{\textbf{1-shot}} & \multicolumn{4}{c}{\textbf{5-shot}}\\
\cmidrule(l){3-6} \cmidrule(l){7-10}
& & Mi-AP & Mi-F1 & Ma-AP & Ma-F1 & Mi-AP & Mi-F1 & Ma-AP & Ma-F1 \\
\midrule
ResNet-50       & - & 18.76 & 23.98 & 22.04 & 22.91  & 24.53 & 28.21 & 27.53 & 28.69  \\ 
ResNet-101      & - & 16.55 & 22.76 & 20.20 & 21.31  & 24.91 & 27.11 & 26.43 & 27.81  \\ 
ViT             & - & 20.19 & 23.97 & 22.44 & 24.29  & 27.40 & 30.29 & 29.45 & 28.67  \\ 
ResNet-50 + ViT & - & 19.26 & 20.45 & 20.35 & 18.41  & 28.86 & 30.48 & 30.24 & 28.79  \\ 
PLA             & Self-learned & 26.14 & 28.09 & 32.40 & 27.86  & 33.72 & 34.31 & 38.85 & 33.15 \\ 
PLA~(GloVe)     & GloVe        & 25.53 & 27.42 & 32.01 & 26.54  & 32.92 & 32.69 & 38.43 & 33.57 \\ 
LaSO            & - & 22.82 & 26.27 & 29.70 & 26.03  & 28.62 & 31.87 & 33.74 & 31.54 \\ 
MAML            & - & 28.32 & 30.02 & 33.76 & 27.95  & 36.35 & 35.43 & 39.31 & 33.82 \\
WGA             & GloVe        & 36.56 & 34.39 & 41.70 & 33.35  & 40.87 & 39.04 & 45.34 & 40.62 \\ 
\midrule
Base model      & GloVe     & 41.18 & 38.48 & 45.26 & 38.56  & 45.29 & 43.32 & 50.30 & 44.14 \\ 
LCM model       & GloVe     & 42.75 & 39.16 & 46.78 & 40.48  & 48.25 & 45.47 & 54.58 & 46.23 \\ 
\midrule
Base model     & mirrorBERT & 42.01 & 38.64 & 46.17 & 38.68  & 47.67 & 44.70 & 52.74 & 46.07  \\ 
LCM model      & mirrorBERT & \textbf{43.32} & \textbf{40.15} & \textbf{48.13} & \textbf{41.05}  
                            & \textbf{48.74} & \textbf{47.13} & \textbf{55.56} & \textbf{47.58} \\ 
\bottomrule
\end{tabular}
\label{nus-wide_fspml}
\end{table*}

\subsection{Baselines}
We compare with LaSO~\cite{laso}, KGGR~\cite{kggr}, \yankun{NLC~\cite{simon2022meta}, BCR~\cite{an2024leveraging}}, which were designed for the ML-FSIC setting. To put the results in context, we also include comparisons with methods tailored for ML-IC. First, following LaSO~\cite{laso}, we attach a standard multi-label classifier to a number of different feature extractors: \emph{ResNet-50}, \emph{ResNet-101}, \emph{ViT-Base}~\cite{vit} and \emph{ResNet-50} + \emph{ViT-Base}. Second, we report results for the recent state-of-the-art CNN-RNN based method PLA~\cite{orderless}, which, to the best of our knowledge, has not previously been evaluated in the ML-FSIC setting. 
Furthermore, we include the seminal meta-learning method MAML~\cite{MAML} designed for FSIC. 
\revision{We also compare our results with our previous work, WGA~\cite{yan2021inferring}. It is worth noting that the base model in this work is an enhanced version of WGA, incorporating additional multi-modal interaction operations, as detailed in Sec.~\ref{proto_cons}.} 
Since we did not have access to the source code of KGGR, \yankun{NLC, and BCR,} we only compared our method against the published results from the original paper, which followed the experimental setting from \cite{laso}. 
For PLA, LaSO, and our method, we used \emph{ResNet-50} as the feature extractor in the main experiments. However, to ensure a fair comparison, we adopt the same feature extractors used in KGGR, LaSO, \yankun{NLC, }\yankun{and BCR,} for comparing our method in the LaSO setting.  

\begin{table}[t]
\caption{Experimental results (mAP in $\%$) on COCO, when following the data and evaluation setting from LaSO \cite{laso}. The best results are indicated by being highlighted in bold font; ``-'' indicates no word embeddings are used.}
\centering
\footnotesize
\begin{tabular}{lcccc}
\toprule
\textbf{Method} & \textbf{\yankun{Backbone}}  & \textbf{\yankun{Word Emb.}} & \textbf{1-shot} & \textbf{5-shot}\\
\midrule
LaSO~\cite{laso}    & \yankun{GoogleNet-v3} & -         & 45.3 & 58.1 \\
KGGR~\cite{kggr}    & \yankun{GoogleNet-v3} & \yankun{GloVe}     & 49.4 & 61.0 \\
\yankun{ProtoNets + NLC~\cite{simon2022meta}}        & \yankun{Conv-4-64}  & -     & \yankun{50.2} & \yankun{60.4} \\
\yankun{RelationNets + NLC~\cite{simon2022meta}}     & \yankun{Conv-4-64}  & -     & \yankun{53.3} & \yankun{60.8} \\
\yankun{LPN + NLC~\cite{simon2022meta}}              & \yankun{Conv-4-64}  & -     & \yankun{56.8} & \yankun{64.8} \\
\yankun{BCR~\cite{an2024leveraging}}        & \yankun{GoogleNet-v3} & -     & \yankun{61.5} & \yankun{70.1} \\
\midrule
\yankun{Base model}   & \multirow{2}{*}{\yankun{Conv-4-64}} & \multirow{2}{*}{\yankun{GloVe}}  & \yankun{59.8} & \yankun{66.7} \\
\yankun{LCM model}    & &  & \yankun{60.3} & \yankun{68.3} \\
\midrule
\yankun{Base model}   & \multirow{2}{*}{\yankun{GoogleNet-v3}}  & \multirow{2}{*}{\yankun{GloVe}}  & \yankun{62.9} & \yankun{71.7} \\
\yankun{LCM model}    & &  & \textbf{\yankun{64.2}} & \textbf{\yankun{73.9}} \\
\bottomrule
\end{tabular}
\label{previous}
\end{table}

\subsection{Experimental Results}
The experimental results for COCO are shown in Tab.~\ref{coco} for the 1-shot and 5-shot settings. 
The results show that our proposed method outperforms the other methods by a substantial margin.
From Tab.~\ref{coco}, it is evident that large models are prone to overfitting, noting that in terms of model size, we have: \emph{ResNet-50} + \emph{ViT-Base} \textgreater \emph{ViT-Base} \textgreater \emph{ResNet-101} \textgreater \emph{ResNet-50}. This is not unexpected given the small number of labeled examples in the 1-shot setting. However, in the 5-shot setting, where more training examples are available, ViT-based models can outperform ResNet-based models. LaSO can improve the ResNet-50 baseline because of its data augmentation strategy. Somewhat surprisingly, PLA performs better than LaSO, which shows that its LSTM component can model label dependencies in a meaningful way. PLA also needs word embeddings, but in the original model, these embeddings are learned from the training data itself. For comparison, we also report results for a variant where these word vectors are initialized using GloVe vectors instead; the results are shown as PLA~(GloVe). As can be seen, using pre-trained word vectors does not lead to a meaningful improvement over the original PLA model and can even degrade the performance on macro metrics, as shown in Tab.~\ref{coco}. Interestingly, MAML~\cite{MAML}, which was designed for the FSIC setting, performs better than the conventional ML-IC approaches, both in the 1-shot and 5-shot settings.
\yankun{The results also show that adding the LCM module improves the result, with the gains being somewhat clearer in the 5-shot setting compared to the 1-shot setting.
}
Please refer to the Appendix for the experimental results for PASCAL VOC, NUS-WIDE, and iMaterialist.
To provide a more thorough evaluation, we conducted additional experiments on both COCO and NUS-WIDE datasets using the data split proposed in FsPML~\cite{zhao2023few}. As shown in Tab.~\ref{coco_fspml} and Tab.~\ref{nus-wide_fspml}, we can again see that our proposed method achieves the best performance.

\yankun{To facilitate a comparison with KGGR, NLC, and BCR, we adopt the 1-shot and 5-shot settings from \cite{laso}. For each method, we utilize the results reported in their respective original papers. As shown in Tab.~\ref{previous}, our method significantly outperforms existing approaches. Notably, our method achieves superior performance even with a simple and shallow Conv-4-64 backbone, whereas both LaSO and KGGR rely on the deeper GoogleNet-v3 backbone. Furthermore, when equipped with GoogleNet-v3, our approach establishes a new state-of-the-art.}

\begin{table}[t]
\caption{\yankun{Ablation study (in $\%$) on components in our method. `Cro. Att.' denotes the channel-wise cross-attention guided by word embedding. `Dyn. Conv.' represents the word-embedding-based dynamic convolution. The best results are highlighted in bold.} }
\centering
\footnotesize
\begin{tabular}{ccccccc}
\toprule
\textbf{\yankun{CM-loss}} & \textbf{\yankun{Cro. Att.}} & \textbf{\yankun{Dyn. Conv.}}  & \textbf{\yankun{LCM}} & \textbf{\yankun{1-shot}} & \textbf{\yankun{5-shot}} & \textbf{\yankun{Notes}} \\
\midrule
   &  \yankun{\checkmark} & \yankun{\checkmark} &  & \yankun{44.01} & \yankun{46.59} & \yankun{Base model w/o CM-loss}  \\
\midrule
 \yankun{\checkmark}   &  & \yankun{\checkmark} & &\yankun{44.92} & \yankun{46.85} & \yankun{Base model w/o Cro. Att.} \\
\midrule
 \yankun{\checkmark}   &  \yankun{\checkmark}  &  & &\yankun{45.67} & \yankun{47.03} & \yankun{Base model w/o Dyn. Conv.} \\
\midrule
 \yankun{\checkmark}   &  \yankun{\checkmark}  & \yankun{\checkmark} &  & \yankun{47.02} & \yankun{48.85}  & \yankun{Base model} \\
\midrule
 \yankun{\checkmark}   &  \yankun{\checkmark}  & \yankun{\checkmark} & \yankun{\checkmark} & \textbf{\yankun{48.92}} & \textbf{\yankun{52.79}}  & \yankun{LCM model} \\
\bottomrule
\end{tabular}
\label{tab:abla_comp}
\end{table}

\begin{table}[t]
\caption{\yk{Results (in $\%$) on COCO for using ``vision and language'' model CLIP. The best results are highlighted in bold.}}
\centering
\footnotesize
\begin{tabular}{llccc}
\toprule
\textbf{\yk{Feature Ext.}} & \textbf{\yk{Word Emb.}} & \textbf{\yk{CM Loss}} & \textbf{\yk{1-shot}} & \textbf{\yk{5-shot}}\\
\midrule
\yk{\multirow{2}[3]{*}{ViT-B/32\textsubscript{CLIP}}} & \yk{\multirow{2}[3]{*}{Word\textsubscript{CLIP}}} & & \yk{53.17} & \yk{56.36} \\
\cmidrule(l){3-5} 
                 &  & \yk{\checkmark}  & \yk{\textbf{55.88}} & \yk{\textbf{60.59}} \\             
\bottomrule
\end{tabular}
\label{tab:vit-varias}
\end{table}

\subsection{Ablation Study}
Here we analyze the importance of the main components of our model. All experiments in this section are conducted on the COCO dataset, employing the \emph{ResNet-50} feature extractor and GloVe vectors, unless stated otherwise. 
The results are reported by default in terms of Ma-AP, commonly referred to as mAP~\cite{laso,cole2021multi,durand2019learning}. For additional ablation study results \revision{and qualitative analysis}, please refer to the Appendix.

\subsubsection{\yankun{Effectiveness of Each Component}}
\yankun{We ablate the effectiveness of each component in our method as shown in Tab.~\ref{tab:abla_comp}. When the component of the CM loss is removed, we only use $\mathcal{L}_{\textit{query}}$ for training. Note that the alignment between the word vectors and visual features then has to be learned indirectly, together with the parameters of the overall model.  As indicated in the first row of Tab.~\ref{tab:abla_comp}, the absence of the CM loss significantly diminishes performance. The second to fourth rows illustrate that removing any multi-modal interaction module also leads to a reduction in performance. Conversely, the inclusion of LCM, as shown in the last row, markedly enhances performance, particularly in the 5-shot setting. One may wonder whether the CM loss is still needed when using a vision-and-language model such as CLIP, given that such models have text and visual spaces which are inherently aligned. We analyse this in Table~\ref{tab:vit-varias}, where we show results on the base model that were obtained with the CLIP (Vit-B/32\textsubscript{CLIP}) image encoder together with the word embeddings from the CLIP text encoder (Word\textsubscript{CLIP}). As can be seen, even in this case, using the CM loss leads to substantial performance gains.}

\begin{table}[t]
\caption{\yankun{Results (in $\%$) on COCO for different feature extractors. The best results are highlighted in bold.}}
\centering
\footnotesize
\begin{tabular}{lccccc}
\toprule
\multirow{2}{*}{\textbf{\yankun{Method}}} & \multirow{2}{*}{\textbf{\yankun{Feature Extractor}}} & \multicolumn{2}{c}{\yankun{\textbf{Results}}} & \multicolumn{2}{c}{\yankun{\textbf{Margin of errors}}} \\
\cmidrule(l){3-4} \cmidrule(l){5-6}
&  & \yankun{\textbf{1-shot}} & \yankun{\textbf{5-shot}} & \yankun{\textbf{1-shot}} & \yankun{\textbf{5-shot}} \\
\midrule
\yankun{Base model} &  \multirow{2}{*}{\yankun{ResNet-50}} & \yankun{47.02} & \yankun{48.85} & \yankun{0.031} & \yankun{0.039} \\
\yankun{LCM model}  &                             & \yankun{48.92} & \yankun{52.79} & \yankun{0.038} & \yankun{0.041} \\
\midrule
\yankun{Base model} & \multirow{2}{*}{\yankun{ResNet-50\textsubscript{CLIP}}}   & \yankun{53.83} & \yankun{56.31} & \yankun{0.054} & \yankun{0.038}  \\
\yankun{LCM model}  &                                                   & \yankun{54.88} & \yankun{60.54} & \yankun{0.057} & \yankun{0.046} \\
\midrule
\yankun{Base model} & \multirow{2}{*}{\yankun{ViT-B/32\textsubscript{CLIP}}}     & \yankun{58.70} & \yankun{61.53} & \yankun{0.055} & \yankun{0.048} \\
\yankun{LCM model}  &                                                   & \textbf{\yankun{60.35}} & \textbf{\yankun{65.01}} & \yankun{0.042} & \yankun{0.037}\\
\bottomrule
\end{tabular}
\label{tab:diff_bbs}
\end{table}

\subsubsection{\yankun{Different Feature Extractors}}
\yankun{For the main experiments, we have primarily used ResNet-50~\cite{residual} as the feature extractor (backbone) for convenient comparison with other methods. Tab.~\ref{tab:diff_bbs} shows the impact of using different feature extractors. In addition to the standard ResNet-50, we also explore backbones from foundation models. Specifically, we leverage ResNet-50 and ViT-B/32 from the CLIP\cite{clip} image encoder (denoted as ResNet-50\textsubscript{CLIP} and ViT-B/32\textsubscript{CLIP} respectively). To adapt them to our approach, we remove the final attention layer and keep all grid features for ResNet-50\textsubscript{CLIP} and ViT-B/32\textsubscript{CLIP} respectively to receive 3D local features. The results presented in Tab.~\ref{tab:diff_bbs} demonstrate remarkable improvements by utilizing feature extractors from CLIP. This observation can be attributed to the extensive training of the CLIP model on a large volume of text-image data, resulting in its image encoder possessing a rich visual knowledge that aligns closely with textual features. }

\subsubsection{\yankun{Robustness of Our Approach}}
\yankun{To verify the robustness of our approach, we train our model 10 times with different seeds, and the margin of error is presented by the standard deviation across these diverse runs. The results in Tab.~\ref{tab:diff_bbs} show the robustness of our approach.}

\begin{table}[t]
\caption{Results (in $\%$) on COCO for different word embeddings. The best results are highlighted in bold.}
\centering
\footnotesize
\begin{tabular}{llcc}
\toprule
\textbf{Method} & \textbf{Word Emb.} & \textbf{\yankun{1-shot}} & \textbf{\yankun{5-shot}}\\
\midrule
Base model &  \multirow{2}{*}{FastText}    & \yankun{39.84} & \yankun{41.19} \\
\yankun{LCM model}  &                               & \yankun{40.67} & \yankun{41.55} \\
\midrule
Base model & \multirow{2}{*}{Skip-Gram}    & \yankun{42.68} & \yankun{43.11} \\
\yankun{LCM model}  &                               & \yankun{43.56} & \yankun{44.87} \\
\midrule
Base model & \multirow{2}{*}{GloVe}        & \yankun{47.02} & \yankun{48.85} \\
\yankun{LCM model}  &                               & \yankun{48.92} & \yankun{52.79} \\
\midrule
\yankun{Base model} & \multirow{2}{*}{\yankun{Word\textsubscript{CLIP}}}   & \yankun{45.34} & \yankun{48.08} \\
\yankun{LCM model}  &                                             & \yankun{46.67} & \yankun{49.32} \\
\midrule
\yankun{Base model} & \multirow{2}{*}{\yankun{BERT\textsubscript{{avg}}}}   & \yankun{40.47} & \yankun{43.41} \\
\yankun{LCM model}  &                                              & \yankun{42.86} & \yankun{44.60} \\
\midrule
\yankun{Base model} & \multirow{2}{*}{\yankun{mirrorWIC}}      & \yankun{42.54} & \yankun{46.16} \\
\yankun{LCM model}  &                                 & \yankun{45.32} & \yankun{48.07} \\
\midrule
\yankun{Base model} & \multirow{2}{*}{\yankun{BERT\textsubscript{biencoder}}}  & \yankun{47.79} & \yankun{49.05} \\
\yankun{LCM model}  &                                                 & \yankun{49.30} & \yankun{52.61} \\
\midrule
\yankun{Base model} & \multirow{2}{*}{\yankun{BERT\textsubscript{concn}}}  & \yankun{49.29} & \yankun{51.25} \\
\yankun{LCM model}  &                                             & \yankun{51.27} & \yankun{54.17} \\
\midrule
\yankun{Base model} & \multirow{2}{*}{\yankun{mirrorBERT}}     & \yankun{49.49} & \yankun{51.49} \\
\yankun{LCM model}  &                                 & \textbf{\yankun{51.55}} & \textbf{\yankun{54.98}} \\
\bottomrule
\end{tabular}
\label{tab:diff_embeds}
\end{table}

\subsubsection{Word Embeddings}\label{secWordEmbeddings}
Tab.~\ref{tab:diff_embeds} compares the results we obtained with different word embeddings: FastText \cite{bojanowski-etal-2017-enriching}, Word2Vec \cite{mikolov-etal-2013-linguistic}, GloVe \cite{glove}, BERT \cite{DBLP:conf/naacl/DevlinCLT19}, and \yankun{word vectors from \steven{the} CLIP text encoder.}
To obtain pre-trained label embeddings from BERT, we considered a number of different strategies, as explained in Sec.~\ref{exp:imple-details}. In comparison to the image encoder of CLIP, as demonstrated in Tab.~\ref{tab:diff_bbs}, the word vectors produced by the text encoder of CLIP (obtained as explained in Section~\ref{exp:imple-details}) exhibit weaker performance than GloVe and some of the BERT strategies.

\begin{figure}
\centering
\subfloat[\yankun{Results (in $\%$) on COCO for varying numbers of attention heads when using the base model.}]{
	\centering
	\includegraphics[width=0.40\textwidth]{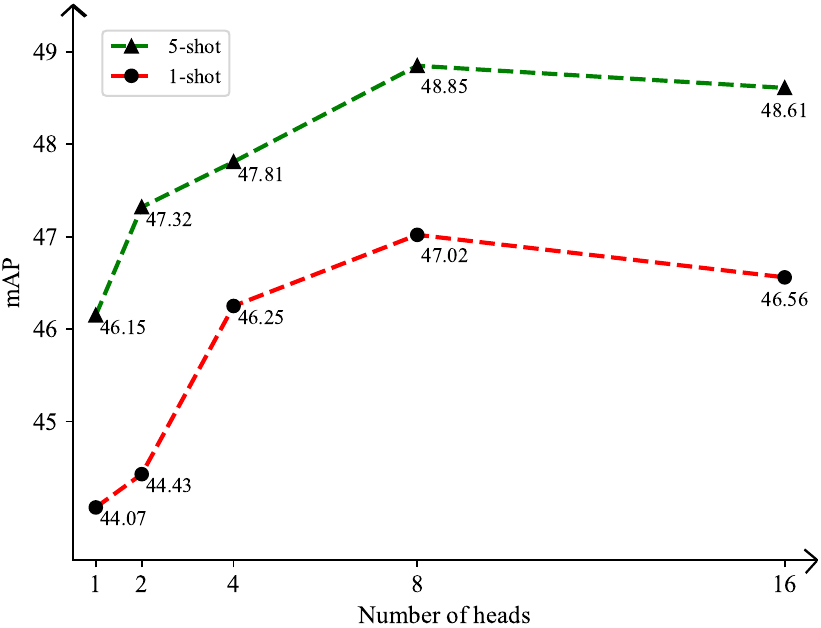}
	\label{fig:num-of-heads}
}
\quad
\subfloat[\yankun{Results (in $\%$) on COCO for different values of the $\theta$ threshold when using the LCM model.}]{
	\centering
	\includegraphics[width=0.40\textwidth]{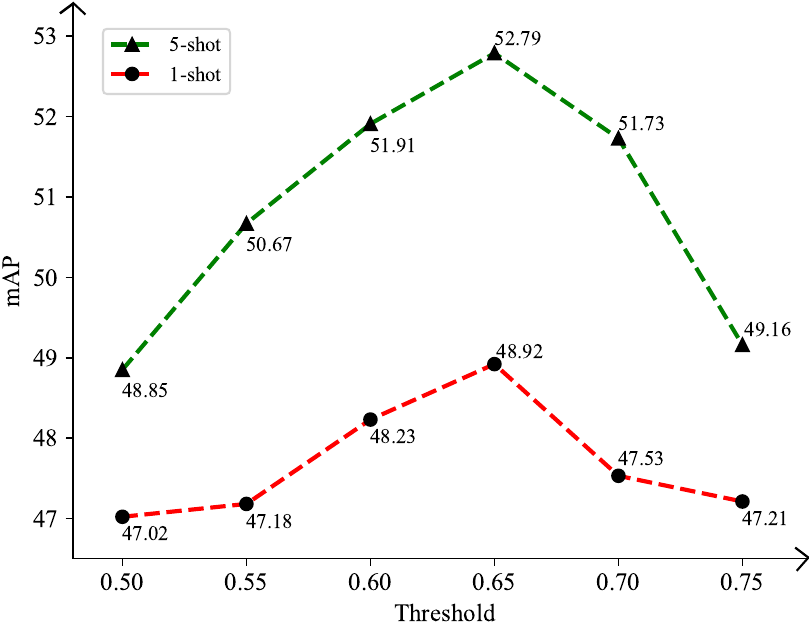}
	\label{fig_thres}
} 
\caption{\yankun{Ablation study for the number of attention heads and threshold $\theta$.}}
\label{fig:num_theta}
\vspace{-2mm}
\end{figure}

\subsubsection{Number of Attention Heads}
\yankun{In Fig.~\ref{fig:num-of-heads},}
we analyze the importance of using multiple attention heads. \yankun{It is important to note that adjustments to the number of attention heads necessitate corresponding changes in the division of visual feature channels. However, despite these changes, the dimensionality of the final prototypes remains constant.}
The results in \yankun{Fig.~\ref{fig:num-of-heads}} show that with fewer than eight attention heads, the performance is notably lower, while there is no obvious benefit to increasing the number of attention heads beyond eight.

\subsubsection{\yankun{LCM Threshold}} 
\yankun{The hyper-parameter $\theta$ controls the threshold for selecting local features based on the learned importance weights. Fig.~\ref{fig_thres} shows the impact of $\theta$ for our LCM model on the COCO dataset.
Note that $\theta = 0.5$ corresponds to the base model since $0.5$ is the minimum value that can be obtained after applying the sigmoid activation function to a non-negative input. As shown, the LCM model achieves the best performance in both 1-shot and 5-shot settings when $\theta = 0.65$. }

\begin{table}[h]
\caption{\revision{Zero-shot results (in $\%$) on COCO of various different word embeddings. The best result is highlighted in bold.}}
\centering
\footnotesize
\begin{tabular}{lcc}
\toprule
\textbf{Word Embedding}  &\textbf{mAP}\\
\midrule
GloVe                    &   18.79   \\
Word\textsubscript{CLIP} &   18.02   \\
mirrorBERT               &   \textbf{23.65}  \\
\bottomrule
\end{tabular}
\label{tab:zero-shot}
\end{table}

\subsection{\revision{Discussion}}
\subsubsection{\revision{Potential for Zero-shot \stevenNew{Classification}}}
\stevenNew{While our method was not designed for the zero-shot setting, the fact that word embeddings are aligned with visual features in the joint space means that they can be used as noisy prototypes, which could in principle be used for zero-shot classification. We test the effectiveness of this approach for} 
three types of word embeddings: the conventional GloVe embeddings~\cite{glove}, Word\textsubscript{CLIP} from the vision-language model CLIP~\cite{clip}, and mirrorBERT~\cite{mirrorbert} from the \stevenNew{BERT} language model. As shown in Tab.~\ref{tab:zero-shot}, GloVe is slightly better than Word\textsubscript{CLIP}. Notably, mirrorBERT significantly outperforms both, with improvements of 4.86 and 5.63 mAP scores over GloVe and Word\textsubscript{CLIP}, respectively. 
These results suggest that embeddings from language models, when aligned with visual features, may be more effective for multi-label zero-shot learning (ML-ZSL) than other label embedding strategies. While these findings are clearly preliminary, we hope these insights can guide the selection of word embeddings in future ML-ZSL methods.

\subsubsection{\revision{Potential Future Directions}}
\revision{There are two key limitations in our method. First, \stevenNew{using off-the-shelf word embeddings for representing labels may not be the optimal strategy. For instance, some approaches for inferring word embeddings from language models are based on contextualised representations, meaning that suitable mentions of these words in a corpus have to be selected. Previous work has shown that selecting the most informative mentions can have an important impact on the quality of the resulting embeddings \cite{DBLP:conf/lrec/WangBAS22}. One promising avenue might thus be to study strategies for selecting those mentions that are most relevant for vision tasks. Exploring the use of Large Language Models for obtaining more informative embeddings also remains as a promising avenue.}
Second, our multi-modal cross-interaction module includes two operations between word embeddings and visual features: channel-wised cross-attention guided by word embedding and word embedding-based dynamic convolution. There may be more effective multi-modal interaction operations to explore. 
}

\begin{figure}[t]
\centering
\includegraphics[width=0.80\columnwidth]{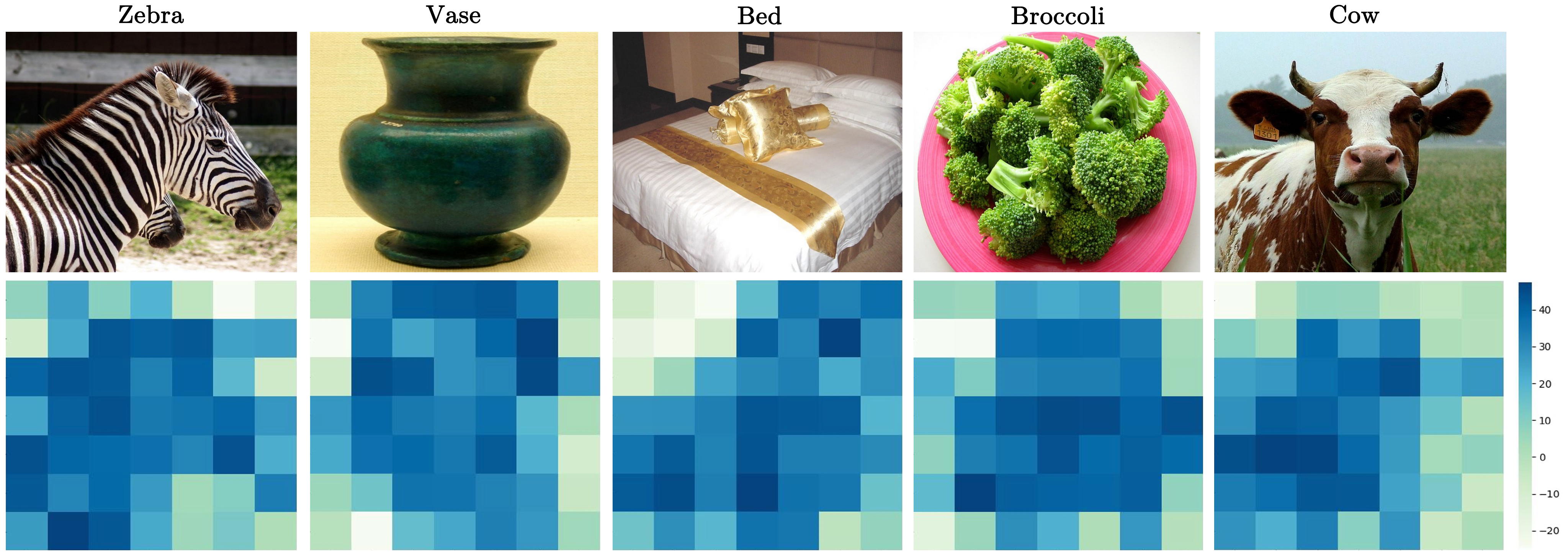} 
\caption{\revision{Visualization of attention weights among local features used to construct label prototypes for large objects. Darker areas indicate higher importance of the corresponding local features.}}
\label{fig:vis_largeobject}
\vspace{-2mm}
\end{figure}

\subsubsection{\revision{More Than Local Objects}}
\revision{As illustrated in Fig.~\ref{fig:vis_largeobject}, our model can effectively identify relevant image regions, even when the label pertains to the content of the whole image. This highlights that our method is not constrained to local objects within an image.}

\section{Conclusion}
We introduced a novel metric-based method for multi-label few-shot image classification. The main challenge, in this setting, comes from the fact that only \revision{a subset of} the local features \revision{from} support images \revision{correspond to the given label. }To construct informative prototypes, \revision{it is essential to estimate which local features are most relevant.} With limited support images, we argue that incorporating prior knowledge can effectively address this challenge.
In particular, we have proposed to use word embeddings to obtain noisy prototypes, which are then processed by a multi-modal cross-interaction mechanism. This mechanism aims to construct the final prototypes by leveraging multiple interactions between visual features and word embeddings.
To further alleviate the influence of irrelevant regions in the image, we have also proposed a loss change measurement module, which analyzes the gradient of the loss function to identify the most representative local features. 
Our model achieved substantially better results than existing models, both on COCO and newly proposed splits of the PASCAL VOC, NUS-WIDE, and iMaterialist datasets.

\begin{acks}
This research was supported in part by the China Postdoctoral Science Foundation (2023M740079, GZC20230058).
 Zied Bouraoui is supported by ANR-22-CE23-0002 ERIANA. Shoaib Jameel is supported by the NVIDIA Academic Hardware Grant. Steven Schockaert was supported by EPSRC grant EP/V025961/1.
\end{acks}

\bibliographystyle{ACM-Reference-Format}
\bibliography{main}

\newpage
\appendix

\section{Experiments}

\begin{table*}[h]
\centering
\footnotesize
\caption{\yankun{Labels covered by the training, validation, and novel sets for the four considered multi-label few-shot image classification datasets.}}
\begin{tabular}{p{1.0cm}p{2.7cm}p{2.7cm}p{2.7cm}p{2.7cm}}
\toprule
 & \multicolumn{4}{c}{\textbf{\yankun{ML-FSIC Datasets}}} \\ 
 \cmidrule(l){2-5} 
& \textbf{\yankun{COCO}} & \textbf{\yankun{PASCAL VOC}} & \textbf{\yankun{NUS-WIDE}}  & \textbf{\yankun{iMaterialist}}\\
\midrule
\yankun{Training Set:}           & \yankun{\{ \textit{hair drier, clock, wine glass, book, cake, tie, motorcycle, sheep, bottle, giraffe, cell phone, suitcase, remote, bench, mouse, carrot, banana, train, sports ball, toothbrush, fire hydrant, airplane, tv, bus, refrigerator, couch, knife, toilet, elephant, truck, parking meter, car, potted plant, kite, skateboard, orange, horse, cat, tennis racket, bowl, scissors, baseball glove, apple, traffic light, handbag, donut, dog, hot dog, oven, umbrella, sink, pizza} \}}  
                       &  \yankun{\{ \textit{person}, \textit{car}, \textit{chair}, \textit{bottle}, \textit{horse}, \textit{bicycle}, \textit{motorbike}, \textit{diningtable} \}}  
                       & \yankun{\{ \textit{animal, whales, railroad, clouds, person, leaf, elk, frost, map, snow, waterfall, toy, wedding, protest, book, running, cat, airport, harbor, valley, soccer, train, sky, moon, swimmers, window, dog, boats, buildings, plants, road, tattoo, nighttime, castle, dancing, plane, zebra, tiger, earthquake, sports, grass, water, glacier, rainbow, fox, rocks, cow, ocean, surf, mountain, computer, flags} \}}  
                       & \yankun{\{\textit{epaulette, fringe, cape, top\_t-shirt\_sweatshirt, sequin, bag\_wallet, bow, leg warmer, shoe, sleeve, pants, dress, headband\_head-covering\_hair-accessory, tassel, ribbon, pocket, umbrella, neckline}  
                       \}}
                       \\
\yankun{Validation Set:}       & \yankun{\{ \textit{cow, dining table, zebra, sandwich, bear, toaster, person, laptop, bed, teddy bear, baseball bat, skis} \}} 
                      & \yankun{\{ \textit{boat}, \textit{cow}, \textit{train}, \textit{aeroplane}, \textit{bus}, \textit{bird} \}} 
                      & \yankun{\{ \textit{tower, tree, food, sun, town, beach, military, birds, flowers, house, lake, garden} \} }
                      & \yankun{\{\textit{watch}, \textit{lapel}, \textit{shorts}, \textit{bead}, \textit{glasses}, \textit{hat}, \textit{sweater}, \textit{tights\_stockings}, \textit{buckle}, \textit{skirt}, \textit{applique}, \textit{rivet} 
                       \}}
                     \\
\yankun{Novel Set:}            & \yankun{\{ \textit{bicycle, boat, stop sign, bird, backpack, frisbee, snowboard,
    surfboard, cup, fork, spoon, broccoli, chair, keyboard, microwave, vase} \}} 
                      & \yankun{\{ \textit{dog}, \textit{sofa}, \textit{cat}, \textit{potted plant}, \textit{tv monitor}, \textit{sheep} \}} 
                      & \yankun{\{ \textit{bear, statue, bridge, sunset, sign, coral, fish, fire, cars, sand, vehicle, cityscape, police, temple, horses, reflection, street} \}} 
                      & \yankun{\{\textit{sock}, \textit{jumpsuit}, \textit{glove}, \textit{scarf}, \textit{coat}, \textit{belt}, \textit{zipper}, \textit{cardigan}, \textit{jacket}, \textit{hood}, \textit{vest}, \textit{flower}, \textit{tie}, \textit{collar}, \textit{ruffle}, \textit{shirt\_blouse}  
                       \}}
                     \\
\bottomrule
\end{tabular}
\label{table:ml-fsic-datasets}
\end{table*}

\subsection{Datasets}
\yankun{An overview of the labels contained in four datasets, including COCO~\cite{lin2014microsoft}, PASCAL VOC \cite{pascal}, NUS-WIDE~\cite{nus-wide}, and iMaterialist~\cite{guo2019imaterialist}, can be found in Tab.~\ref{table:ml-fsic-datasets}.
}

\subsection{Experimental Results}
The experimental results for PASCAL VOC, NUS-WIDE, and iMaterialist are shown in Tab.~\ref{voc}, Tab.~\ref{nus-wide}, and Tab.~\ref{fashion}, respectively. Our method consistently achieves the best performance.

\begin{table*}[t]
\caption{
Overview of the main results on PASCAL VOC, in terms of micro/macro AP and F1 scores (in $\%$).
The best results are highlighted in bold font.
``-'' indicates that no word embeddings are used.}
\centering
\footnotesize
\begin{tabular}{l c cccccccc}
\toprule
 & \multirow{2}{*}{Word-Emb.} & \multicolumn{4}{c}{\textbf{1-shot}} & \multicolumn{4}{c}{\textbf{5-shot}}\\
\cmidrule(l){3-6} \cmidrule(l){7-10}
& & Mi-AP & Mi-F1 & Ma-AP & Ma-F1                & Mi-AP & Mi-F1 & Ma-AP & Ma-F1 \\
\midrule
ResNet-50   & - & 16.57 & 20.77 & 27.24 & 21.92    & 24.47 & 28.53 & 35.06 & 29.43 \\ 
ResNet-101  & - & 19.23 & 23.86 & 30.11 & 22.17    & 27.07 & 29.37 & 39.01 & 31.26 \\ 
ViT         & - & 16.69 & 17.76 & 27.72 & 17.15    & 31.70 & 37.35 & 40.13 & 36.29  \\ 
ResNet-50 + ViT    & - & 18.71 & 13.70 & 28.76  & 12.86    & 34.69 & 35.83 & 42.73  & 33.99 \\ 
PLA                & Self-learned & 41.98 & 36.81 & 47.29  & 37.83    & 48.09 & 42.01 & 55.54  & 45.40  \\ 
PLA~(GloVe)        & GloVe & 41.31 & 35.84 & 46.54  & 36.55    & 47.50 & 40.60 & 55.26  & 45.33   \\ 
LaSO               & - & 22.12 & 27.02 & 32.50  & 25.20    & 30.61 & 33.64 & 41.89  & 34.96 \\ 
MAML               & - & 40.21 & 35.11 & 44.15  & 36.08    & 45.95 & 39.26 & 50.03  & 42.74 \\
\revision{WGA}   & \revision{GloVe} & \revision{46.28} & \revision{40.19} & \revision{53.26}  & \revision{43.35} 
                 & \revision{51.10}  & \revision{45.03} & \revision{58.27} & \revision{48.20} \\
\midrule
Base model  & GloVe & \yankun{49.92} & \yankun{44.00} & \yankun{56.59} & \yankun{47.05}  & \yankun{54.35} & \yankun{48.21}  & \yankun{61.48} & \yankun{51.21}  \\ 
LCM model   & GloVe & \yankun{50.98} & \yankun{44.18}  & \yankun{58.03} & \yankun{47.31}    & \yankun{56.74} & \yankun{49.08} & \yankun{65.41} & \yankun{52.00} \\ 
\midrule
Base model  & mirrorBERT & \yankun{51.19} & \yankun{44.37}  & \yankun{58.43} & \yankun{47.96} & \yankun{55.36} & \yankun{49.37}  & \yankun{63.33} & \yankun{51.55}  \\ 
LCM model   & mirrorBERT & \textbf{\yankun{53.30}} & \textbf{\yankun{46.05}} & \textbf{\yankun{59.13}} & \textbf{\yankun{49.65}}  
   & \textbf{\yankun{58.30}} & \textbf{\yankun{51.19}} & \textbf{\yankun{66.78}} & \textbf{\yankun{54.01}} \\ 
\bottomrule
\end{tabular}
\label{voc}
\end{table*}

\begin{table*}[t]
\caption{
\yankun{Overview of the main results on NUS-WIDE, in terms of micro/macro AP and F1 scores (in $\%$). The best results are highlighted in bold font. ``-'' indicates that no word embeddings are used.}}
\centering
\footnotesize
\begin{tabular}{l c cccccccc}
\toprule
  & \multirow{2}{*}{Word-Emb.} & \multicolumn{4}{c}{\textbf{1-shot}} & \multicolumn{4}{c}{\textbf{5-shot}}\\
\cmidrule(l){3-6} \cmidrule(l){7-10}
& & Mi-AP & Mi-F1 & Ma-AP & Ma-F1 & Mi-AP & Mi-F1 & Ma-AP & Ma-F1 \\
\midrule
ResNet-50       & - & \yankun{16.46} & \yankun{21.49} & \yankun{20.78} & \yankun{21.77} & \yankun{21.71} & \yankun{25.33} & \yankun{24.89} & \yankun{26.02}\\ 
ResNet-101      & - & \yankun{15.10} & \yankun{20.27} & \yankun{17.90} & \yankun{20.18} & \yankun{22.60} & \yankun{25.82} & \yankun{24.81} & \yankun{26.25} \\ 
ViT             & - & \yankun{17.08} & \yankun{20.76} & \yankun{19.94} & \yankun{21.97} & \yankun{24.85} & \yankun{28.58} & \yankun{27.04} & \yankun{26.57}  \\ 
ResNet-50 + ViT & - & \yankun{16.05} & \yankun{18.92} & \yankun{18.10}  & \yankun{15.13}  & \yankun{26.32} & \yankun{27.84} & \yankun{29.09}  & \yankun{25.15} \\ 
PLA             & Self-learned & \yankun{22.31} & \yankun{23.92} & \yankun{27.85}  & \yankun{24.63} & \yankun{30.32} & \yankun{31.59} & \yankun{35.24}  & \yankun{30.33} \\ 
PLA~(GloVe)     & GloVe & \yankun{22.47} & \yankun{23.86} & \yankun{27.69} & \yankun{24.29}  & \yankun{29.78} & \yankun{30.47} & \yankun{36.01}  & \yankun{31.04} \\ 
LaSO            & - & \yankun{20.27} & \yankun{23.17} & \yankun{25.95}  & \yankun{22.33}  & \yankun{25.35} & \yankun{28.91} & \yankun{29.63}  & \yankun{27.77} \\ 
MAML            & - & \yankun{25.32} & \yankun{26.94} & \yankun{30.09}  & \yankun{24.82}  & \yankun{32.95} & \yankun{32.52} & \yankun{36.17}  & \yankun{30.50} \\ 
\revision{WGA}   & \revision{GloVe} & \revision{32.91} & \revision{31.53} & \revision{38.12}  & \revision{30.71} 
                 & \revision{36.81}  & \revision{35.16} & \revision{41.25} & \revision{35.63} \\
\midrule
Base model & GloVe  & \yankun{37.88} & \yankun{35.31} & \yankun{42.39} & \yankun{35.53} & \yankun{42.17} & \yankun{40.36} & \yankun{47.02} & \yankun{40.80}  \\ 
LCM model  & GloVe  & \yankun{38.77} & \yankun{36.09} & \yankun{44.96} & \yankun{36.08} & \yankun{45.65} & \yankun{42.31} & \yankun{50.91} & \yankun{43.65} \\ 
\midrule
Base model  & mirrorBERT & \yankun{38.85} & \yankun{35.32} & \yankun{43.53} & \yankun{35.89}  & \yankun{44.16} & \yankun{41.85} & \yankun{48.54} & \yankun{42.54}  \\ 
LCM model   & mirrorBERT & \textbf{\yankun{39.64}} & \textbf{\yankun{36.61}} & \textbf{\yankun{45.00}} & \textbf{\yankun{37.30}}  
       & \textbf{\yankun{46.19}} & \textbf{\yankun{43.81}} & \textbf{\yankun{52.58}} & \textbf{\yankun{44.43}} \\ 
\bottomrule
\end{tabular}
\label{nus-wide}
\end{table*}

\begin{table*}[t]
\caption{
\yankun{Overview of the main results on iMaterialist, in terms of micro/macro AP and F1 scores (in $\%$). The best results are highlighted in bold font. ``-'' indicates that no word embeddings are used.}}
\centering
\footnotesize
\begin{tabular}{l c cccccccc}
\toprule
  & \multirow{2}{*}{\yankun{Word-Emb.}} & \multicolumn{4}{c}{\yankun{\textbf{1-shot}}} & \multicolumn{4}{c}{\yankun{\textbf{5-shot}}}\\
\cmidrule(l){3-6} \cmidrule(l){7-10}
& & \yankun{Mi-AP} & \yankun{Mi-F1} & \yankun{Ma-AP} & \yankun{Ma-F1}   
& \yankun{Mi-AP} & \yankun{Mi-F1} & \yankun{Ma-AP} & \yankun{Ma-F1} \\
\midrule
\yankun{ResNet-50}   & \yankun{-} & \yankun{23.58} & \yankun{27.02} & \yankun{22.16} & \yankun{19.33}    
                     & \yankun{34.90} & \yankun{31.84} & \yankun{30.06} & \yankun{25.10} \\ 
\yankun{ResNet-101}  & \yankun{-} & \yankun{16.71} & \yankun{26.35} & \yankun{21.76} & \yankun{21.45}   
                     & \yankun{38.79} & \yankun{32.89} & \yankun{34.39} & \yankun{28.32} \\ 
\yankun{ViT}         & \yankun{-} & \yankun{23.39} & \yankun{29.45} & \yankun{24.20} & \yankun{15.54}    
                     & \yankun{38.98} & \yankun{37.18} & \yankun{35.44} & \yankun{28.20}  \\ 
\yankun{ResNet-50 + ViT} & \yankun{-} & \yankun{24.32} & \yankun{30.50} & \yankun{25.27}  & \yankun{17.13}    
                         & \yankun{40.46} & \yankun{38.96} & \yankun{35.73}  & \yankun{28.44} \\ 
\yankun{PLA}             & \yankun{Self-learned} & \yankun{31.61} & \yankun{27.82} & \yankun{36.41} & \yankun{27.65}    
                         & \yankun{41.69} & \yankun{30.78} & \yankun{37.08}  & \yankun{30.65} \\ 
\yankun{PLA~(GloVe)}     & \yankun{GloVe} & \yankun{30.43} & \yankun{25.94} & \yankun{34.97}  & \yankun{27.46}    
                         & \yankun{40.46} & \yankun{29.91} & \yankun{36.49}  & \yankun{30.09} \\ 
\yankun{LaSO}            & \yankun{-} & \yankun{29.69} & \yankun{25.31} & \yankun{30.51}  & \yankun{24.82}    
                         & \yankun{38.78} & \yankun{29.25} & \yankun{35.87}  & \yankun{27.82} \\ 
\yankun{MAML}            & \yankun{-} & \yankun{31.42} & \yankun{29.40} & \yankun{37.96}  & \yankun{27.32}   
                         & \yankun{41.70} & \yankun{31.10} & \yankun{39.89}  & \yankun{30.17} \\
\revision{WGA}   & \revision{GloVe} & \revision{41.01} & \revision{38.64} & \revision{46.37} & \revision{34.74} 
                                    & \revision{52.13} & \revision{41.48} & \revision{47.26} & \revision{38.32} \\
\midrule
\yankun{Base model}  & \yankun{GloVe} & \yankun{45.90} & \yankun{43.45}  & \yankun{51.23} & \yankun{40.06}   
                                & \yankun{57.96} & \yankun{46.60}  & \yankun{52.57} & \yankun{43.49}  \\ 
\yankun{LCM model}   & \yankun{GloVe} & \yankun{47.93} & \yankun{44.13} & \yankun{53.09} & \yankun{41.07}  
                                & \yankun{61.50} & \yankun{\textbf{48.97}} & \yankun{56.74} & \yankun{45.15} \\ 
\midrule
\yankun{Base model}  & \yankun{mirrorBERT} & \yankun{48.26} & \yankun{43.80}  & \yankun{54.25} & \yankun{42.13}   
                                & \yankun{59.83} & \yankun{47.07}  & \yankun{54.96} & \yankun{44.89}  \\ 
\yankun{LCM model}   & \yankun{mirrorBERT} & \textbf{\yankun{50.09}} & \textbf{\yankun{44.71}} & \textbf{\yankun{55.27}} & \textbf{\yankun{42.25}}  
      & \yankun{\textbf{62.43}} & \yankun{48.45} & \yankun{\textbf{57.62}} & \yankun{\textbf{45.18}} \\ 
\bottomrule
\end{tabular}
\label{fashion}
\end{table*}

\subsection{Ablation Study}

%

\begin{table}[t]
\caption{Results (in $\%$) on COCO for different \yankun{attention mechanism}. The best results are highlighted in bold.}
\centering
\footnotesize
\begin{tabular}{lccc}
\toprule
\textbf{Method} & \textbf{Difference} & \textbf{\yankun{1-shot}} & \textbf{\yankun{5-shot}}\\
\midrule
Base model & \multirow{2}{*}{Simple attention}     & \yankun{41.42} & \yankun{43.25} \\
LCM model &                                            & \yankun{41.83} & \yankun{44.17} \\
\midrule
Base model & \multirow{2}{*}{\yankun{Standard attention}}     & \yankun{45.84} & \yankun{46.91} \\
LCM model &                                                      & \yankun{47.00} & \yankun{49.45} \\
\midrule
Base model & \multirow{2}{*}{Attention with global features}     & \yankun{42.68} & \yankun{45.97} \\
LCM model &                                                      & \yankun{43.11} & \yankun{47.60} \\
\midrule
Base model & \multirow{2}{*}{Low-rank bilinear pooling}    & \yankun{38.82} & \yankun{41.00} \\
LCM model &                                                & \yankun{39.29} & \yankun{42.71} \\
\midrule
Base model & \multirow{2}{*}{-}      & \yankun{47.02} & \yankun{48.85} \\
LCM model &                          & \textbf{\yankun{48.92}} & \textbf{\yankun{52.79}} \\
\bottomrule
\end{tabular}
\label{tab:diff_varias}
\end{table}


\subsubsection{Importance of the Attention Mechanism}
To evaluate the effect of the proposed attention mechanism, in Tab.~\ref{tab:diff_varias}, we report results for a variant that relies on a simpler mechanism for generating prototypes (shown as \emph{Simple attention}). In particular, in this variant, we generate the \yankun{$\mathbf{p}_{c}^{att}$} of a label $c$ by taking a weighted average of the global features of the support images which have that label. The weights are obtained by computing the cosine similarity between the vectors $\widehat{\mathbf{w}_c}$ and the feature vectors $\widehat{\mathbf{f}^I_{\textit{glo}}}$, multiplying these cosine similarities with the scalar $\lambda$, and feeding the resulting values to a softmax layer. As shown in Tab.~\ref{tab:diff_varias}, the performance degradation in this simplified setting is clearly evident.
\yankun{We also present results for a variant employing the standard attention mechanism, denoted as \emph{Standard Attention}. In this variant, the inputs of \emph{value} and \emph{key} are generated by processing visual features through fully connected layers, rather than splitting them evenly along the channel dimension.}
\yankun{Additionally, we report results when local features are used instead of global features within our cross-attention mechanism (\textit{Attention with global features}).}
Finally, we experimented with a variant that learns visual prototypes in the same way as in KGGR, where the weights of local features are generated by a low-rank bilinear pooling method followed by a fully connected layer with softmax activation (\textit{Low-rank bilinear pooling}).
The results in Tab.~\ref{tab:diff_varias} show that all of these variants result in lower performance.

\begin{table}[t]
\caption{Results (in $\%$) on COCO for different ways to select representative local features for the LCM model based on the 5-shot setting. The best results are highlighted in bold.}
\centering
\footnotesize
\begin{tabular}{lcc}
\toprule
\textbf{Method} & \textbf{Ma-AP} & \textbf{Mi-AP}\\
\midrule
Base model                           & \yankun{48.85} & \yankun{42.84}\\
Using $\rho_{j,k}$                   & \yankun{51.46} & \yankun{45.18} \\
Directly using $g(\cdot)$            & \yankun{50.16} & \yankun{43.55} \\
Taylor approximation                 & \yankun{51.73} & \yankun{45.63} \\
LCM model                            & \textbf{\yankun{52.79}} & \textbf{\yankun{46.35}} \\
\bottomrule
\end{tabular}
\label{imp_scores}
\end{table}

\subsubsection{\yankun{Selection of Local Features}}
\yankun{There are a number of ways in which the LCM module can be simplified. 
First, we could use the importance weights $\rho_{j,k}$ directly, i.e.\ we could select those local features whose $\rho_{j,k}$-value is above a certain threshold. Next, the implementation of loss change measurement itself could also be simplified. For instance, we could evaluate the loss change $g(\cdot)$ directly using Eq.~\ref{loss_change}. A final possibility we consider is to use the Taylor approximation, i.e.\ to use Eq.~\ref{eqLCMapproximation}, but without the momentum-based strategy. These different variants are compared with the full LCM model in Tab.~\ref{imp_scores}. 
In addition, we include results of the base model in the table for comparison. Notably, all variants exhibit superior performance compared to the base model, with the proposed LCM model attaining the best overall results.}

\subsubsection{\yankun{Local Features for Query Images}}
\yankun{In our method, we initially compare the global features of query images with prototypes. Additionally, we experimented with 
a variant of the base model where the local features of the query image are matched against those of the prototypes. In this variant, the similarity score is derived from the sum of the top-$k$ values, with $k=12$, chosen based on the validation split. This variant achieves mAP scores of 47.54 and 49.11 in the 1-shot and 5-shot settings, respectively. While this represents a slight improvement over our base model (47.02/48.85), it incurs a significantly higher computational cost.}

\begin{figure}[t]
\centering
\includegraphics[width=0.95\textwidth]{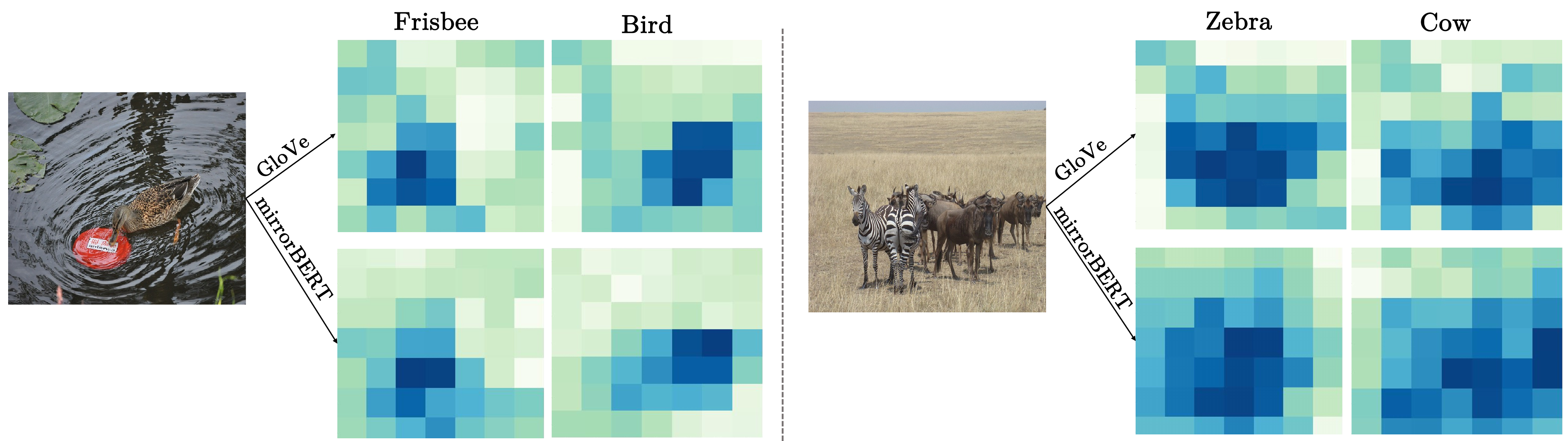} 
\caption{\yankun{Visualization of the attention weights among local features for constructing label prototypes, \steven{using} GloVe and mirrorBERT word vectors, across various categories. The \steven{color} intensity indicates the relative importance of the corresponding local feature.}}
\label{fig_vis}
\end{figure}

\subsubsection{Visualization of the Attention Mechanism}
Fig.~\ref{fig_vis} illustrates which regions the proposed attention mechanism focuses on \yankun{according to different categories for two choices of the word vectors}. We take the sum of attention weights across all attention heads as the visualization scores of the local features. As illustrated in the first example of Fig.~\ref{fig_vis}, the model is often successful in identifying the most relevant image region \yankun{for different categories and the attention weights are indeed label-specific}. 
In this example, the model correctly selects the frisbee or the duck depending on the selected label. This is even though no images with these labels were present in the training data. On the other hand, as the last example shows, for labels that are semantically closely related, such as \emph{zebra} and \emph{cow} in this case, word vectors are not sufficiently informative. 
For both labels, the model correctly selects the group of animals, but it fails to make a finer selection.

\begin{figure}[t]
\centering
\includegraphics[width=0.87\columnwidth]{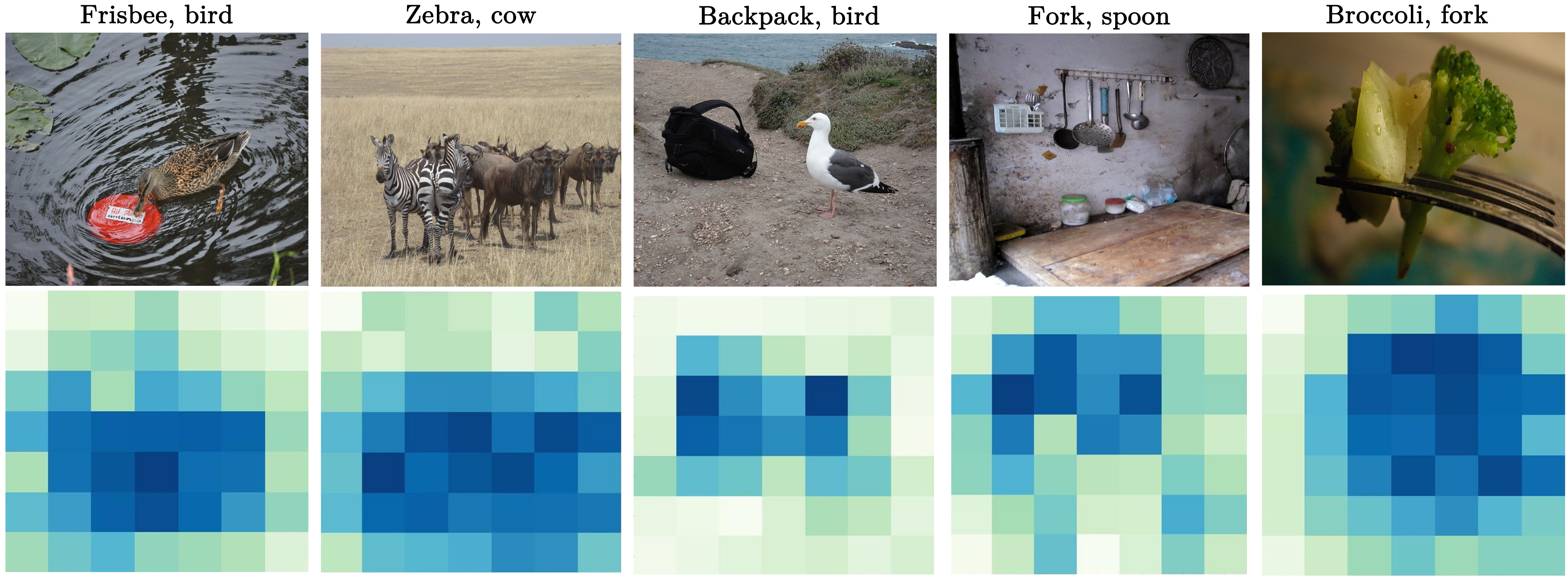} 
\caption{\yankun{Visualization of the distribution of importance scores among local features in the LCM module. The more important the local feature, the darker the corresponding visualization.}}
\label{fig_vis2}
\end{figure}

\subsubsection{\yankun{Visualization of LCM Importance Weights}}
\yankun{Fig.~\ref{fig_vis2} visualizes the importance of the different local features, as predicted by the LCM module. The darker the color, the more important the local features are considered to be. Most background local features are given low scores. In fact, many of these features will have a score below the threshold $\theta$ and will thus be completely ignored when aggregating prototypes.}

\end{document}